\documentclass[10pt,journal,compsoc]{IEEEtran}
\usepackage{amsmath,amsfonts}
\usepackage{textcomp}
\usepackage{url}
\usepackage{verbatim}
\usepackage{graphicx}
\usepackage{cite}
\usepackage{subfigure}
\usepackage{amsmath,amssymb,amsfonts}
\usepackage{algorithmic}
\usepackage[ruled,linesnumbered,lined,boxed,commentsnumbered]{algorithm2e}
\usepackage{multirow}
\usepackage{amssymb}
\usepackage{pifont}
\usepackage{xcolor}
\usepackage{tabularx}
\usepackage{color}
\usepackage{makecell}
\usepackage{pifont}

\begin{document}

\title{DVFO: Learning-Based DVFS for Energy-Efficient Edge-Cloud Collaborative Inference}

\author{Ziyang~Zhang,~\IEEEmembership{Student Member,~IEEE,}
        Yang~Zhao,~\IEEEmembership{Senior Member, IEEE,}
        Huan~Li,\\~\IEEEmembership{Senior Member, IEEE,}
        Changyao~Lin,
        and~Jie~Liu,~\IEEEmembership{Fellow,~IEEE}
\IEEEcompsocitemizethanks{\IEEEcompsocthanksitem Ziyang Zhang and Changyao Lin are with the School of Science and Technology, Harbin Institute of Technology, Harbin, Heilongjiang 150006, China.
E-mail: \{zhangzy,lincy\}@stu.hit.edu.cn \protect
\IEEEcompsocthanksitem Yang Zhao, Huan Li and Jie Liu are with the International Research Institute for Artificial Intelligence, Harbin Institute of Technology, Shenzhen, Guangdong 518071, China.
E-mail: \{yang.zhao,huanli,jieliu\}@hit.edu.cn.}
\thanks{Manuscript received June 28, 2023; revised September 26, 2023. \\
This work is partly supported by the National Key R\&D Program of China under Grant No. 2021ZD0110905, and An Open Competition Project of Heilongjiang Province, China, on Research and Application of Key Technologies for Intelligent Farming Decision Platform, under Grant No. 2021ZXJ05A03.\\
(Corresponding author: Jie Liu.)}}

\IEEEtitleabstractindextext{%
\begin{abstract}
Due to limited resources on edge and different characteristics of deep neural network (DNN) models, it is a big challenge to optimize DNN inference performance in terms of energy consumption and end-to-end latency on edge devices. 
In addition to the dynamic voltage frequency scaling (DVFS) technique, the edge-cloud architecture provides a collaborative approach for efficient DNN inference.
However, current edge-cloud collaborative inference methods have not optimized various compute resources on edge devices. 
Thus, we propose DVFO, a novel DVFS-enabled edge-cloud collaborative inference framework, which co-optimizes DVFS and offloading parameters via deep reinforcement learning (DRL). 
Specifically, DVFO automatically co-optimizes 
1) the CPU, GPU and memory frequencies of edge devices, and 
2) the feature maps to be offloaded to cloud servers. 
In addition, it leverages a \emph{thinking-while-moving} concurrent mechanism to accelerate the DRL learning process, and a \emph{spatial-channel attention} mechanism to extract DNN feature maps of secondary importance for workload offloading. 
This approach improves inference performance for different DNN models under various edge-cloud network conditions.
Extensive evaluations using two datasets and six widely-deployed DNN models on three heterogeneous edge devices show that DVFO significantly reduces the energy consumption by 33\% on average, compared to state-of-the-art schemes. 
Moreover, DVFO achieves up to 28.6\%$\sim$59.1\% end-to-end latency reduction, while maintaining accuracy within 1\% loss on average.
\end{abstract}

}

\maketitle
\IEEEpeerreviewmaketitle

\section{Introduction}\label{sec:introduction}
\IEEEPARstart {A}{s} the development of edge computing and deep learning techniques, edge devices equipped with internet of things (IoT) connectivity and hardware accelerators (e.g., GPUs) are becoming capable of executing deep neural network (DNN) in real-time for many edge intelligence~\cite{chen2019deep} applications, such as scene perception in autonomous driving~\cite{jang2020r}, defect detection in industry~\cite{gao2020real} and face recognition in smartphones~\cite{boutros2022elasticface}, etc.
However, compared to cloud servers, edge devices have fewer compute resources and more stringent power consumption requirements, thus it is more challenging to optimize DNN inference performance in terms of energy consumption and end-to-end latency on edge devices~\cite{han2022microsecond}.
\begin{figure}[htbp]
\vspace{-0.2cm}
\large
\centerline{\includegraphics[width=0.85\linewidth]{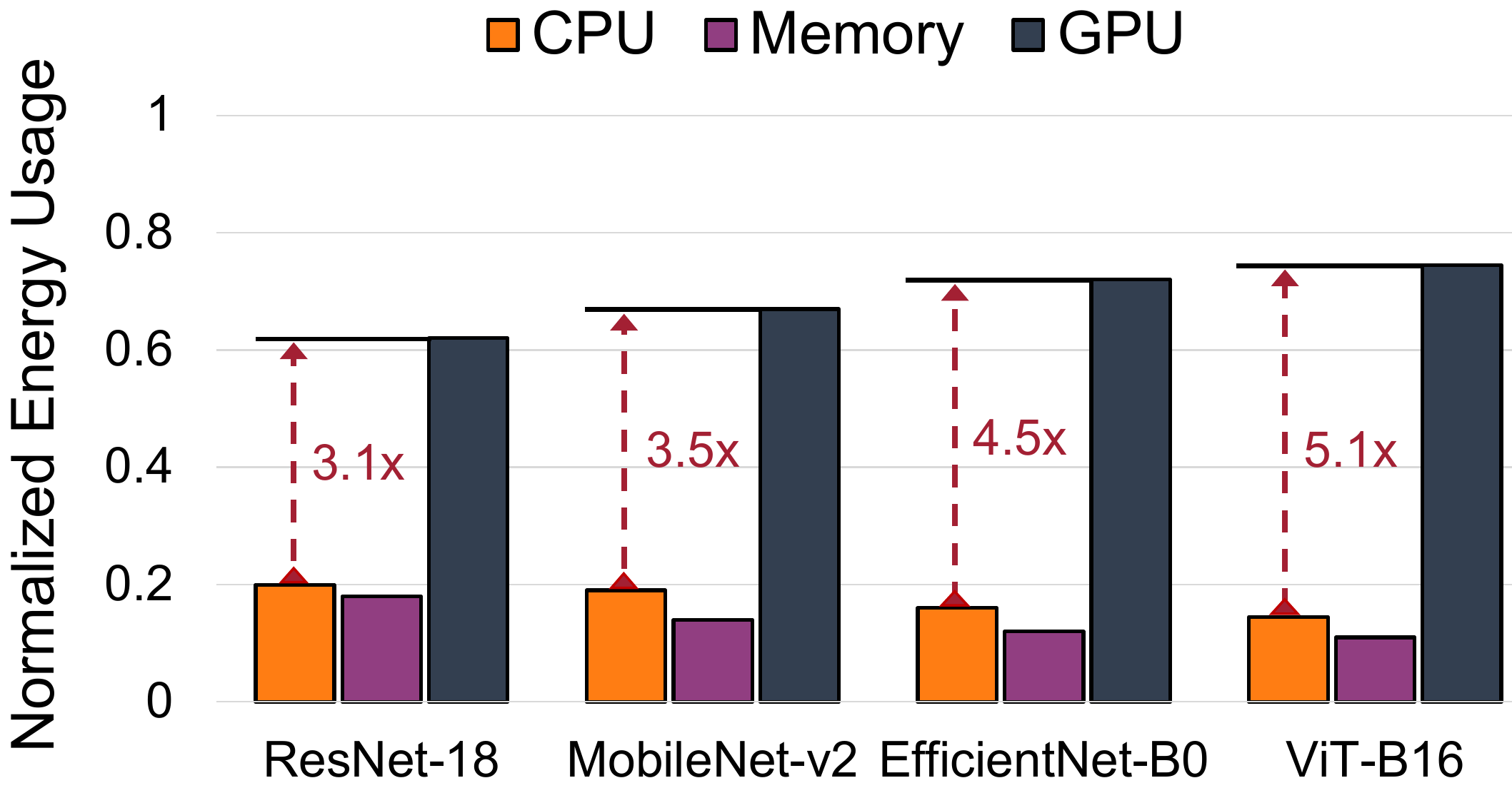}}
\caption{Energy usage of CPU, GPU and memory for four DNN inference models with CIFAR-100~\cite{krizhevsky2009learning} dataset, measured on NVIDIA Xavier NX. We set the batch size to 1.}
\label{fig:energy}
\end{figure}

To achieve efficient DNN inference on resource-constrained edge devices, it is a promising approach to reduces the end-to-end latency or energy consumption of edge devices via various techniques such as dynamic voltage frequency scaling (DVFS)~\cite{wang2022energy,nabavinejad2022coordinated}, and edge-cloud collaborative inference~\cite{laskaridis2020spinn,huang2022real}. DVFS is a low-power technology that dynamically adjusts the voltage and frequency according to energy consumption. 
Prior work~\cite{chen2022quality} has proposed a series of deep reinforcement learning-based DVFS techniques to reduce energy consumption. 
However, DVFS reduces energy consumption by increasing end-to-end latency, which we illustrate and discuss in Section~\ref{motivation}. In addition, none of the existing methods above considers the edge-cloud collaboration paradigm. 
The edge-cloud collaborative inference offloads partial DNN feature maps from edge devices to cloud servers, with edge devices inferring partial DNN, cloud servers executing the rest, and small neural networks to fuse them to obtain the final inference results~\cite{yao2020deep}.
To avoid network bottlenecks to achieve offloading DNN feature maps efficiently, prior work utilizes explainable AI~\cite{huang2022real} and compressed sensing~\cite{yao2020deep} to compress feature maps. 
However, the expensive runtime overhead of these schemes still impairs DNN inference real-time performance. 

Combining DVFS and edge-cloud collaboration, prior work~\cite{panda2022energy} proposes a data offloading scheme, namely DRLDO, which uses deep reinforcement learning together with DVFS to reduce energy consumption. 
However, DRLDO only considers CPU core voltage and frequency in DVFS, without including the GPU and memory resources. In addition, it does not consider performance bottlenecks of various DNN models. 
Recent benchmarks reveal that GPUs are responsible for around 70\% of the total energy consumption during DNN training~\cite{dodge2022measuring}. 
As shown in Fig.\ref{fig:energy}, we perform experiments and show that during DNN inference phase, GPUs also consume more energy than CPUs for all the DNN models that we have investigated. 
We report the normalized energy usage of different compute units including CPU, GPU, and memory, when executing four DNN models with CIFAR-100~\cite{krizhevsky2009learning} dataset on an NVIDIA Xavier NX edge device. 
The result shows that the energy consumption of the GPU is 3.1$\times$ to 3.5$\times$ that of the CPU, indicating that GPU dominates DNN inference. 
It can also be observed that since DNN inference accesses memory frequently, the energy consumption of the memory is not negligible. 
In addition, as shown in Fig.~\ref{fig:observation}, the performance of different DNN models has diminishing returns as hardware frequencies increase. 
Learning DNN model behaviors on different edge devices can further improve inference performance and energy efficiency.
All these observations motivate us to incorporate CPU, GPU and memory resources in DVFS, and utilize feature maps offloading for DNN inference on edge devices. 

Table~\ref{related} provides a comparison of key features of DVFO with four dimensions of DVFO to related work, including DVFS technology and edge-cloud collaborative inference.
DVFS technology enables on-device DNN inference with lower energy consumption.
While DRLDO~\cite{panda2022energy}, CARTAD~\cite{liu2021cartad} and QL-HDS~\cite{zhang2017energy} have achieved energy-efficient inference on multi-core CPU systems using DVFS technology, they did not consider edge devices with CPU-GPU heterogeneous processors, which are crucial for GPU-dominated energy-efficient on-device inference.
DeepCOD~\cite{yao2020deep} and AgileNN~\cite{huang2022real} compressed the offloaded DNN feature maps, but the compression overhead is not negligible.
Since most of the works mentioned above do not combine DVFS with edge-cloud collaborative inference, in this paper we showcase how to achieve low latency and energy consumption using learning-based DVFS in an edge-cloud collaboration framework. 
\begin{table}
\scriptsize
\setlength{\abovecaptionskip}{0pt}
\setlength{\belowcaptionskip}{0pt}
    \caption{Comparison of key features of DVFO with prior work}
    \label{related}
    \centering
    \begin{tabular}{ccccc} \hline
    	\textbf{\makecell[c]{Service \\ Framework}}  & \textbf{\makecell[c]{Enable \\ DVFS}} & \textbf{\makecell[c]{Collaborative \\ Inference}} & \textbf{\makecell[c]{Data \\ Compression}} & \textbf{\makecell[c]{Enable \\ GPU device}} \\ \hline
     
    	DRLDO~\cite{panda2022energy}  & \textcolor{teal}{\checkmark} & \textcolor{teal}{\checkmark} & \textcolor{purple}{\ding{55}}  & \textcolor{purple}{\ding{55}} \\ 

    	CARTAD~\cite{liu2021cartad} & \textcolor{teal}{\checkmark} & \textcolor{purple}{\ding{55}}  & \textcolor{purple}{\ding{55}}  & \textcolor{purple}{\ding{55}} \\
     
        QL-HDS~\cite{zhang2017energy} & \textcolor{teal}{\checkmark} & \textcolor{purple}{\ding{55}} & \textcolor{purple}{\ding{55}}  & \textcolor{purple}{\ding{55}} \\  

        AppealNet~\cite{li2021appealnet} & \textcolor{purple}{\ding{55}} & \textcolor{teal}{\checkmark} & \textcolor{purple}{\ding{55}}  & \textcolor{teal}{\checkmark} \\
        
        DeepCOD~\cite{yao2020deep} & \textcolor{purple}{\ding{55}} & \textcolor{teal}{\checkmark} & \textcolor{teal}{\checkmark}  & \textcolor{teal}{\checkmark} \\ 

        AgileNN~\cite{huang2022real} & \textcolor{purple}{\ding{55}} & \textcolor{teal}{\checkmark}  & \textcolor{teal}{\checkmark}  & \textcolor{teal}{\checkmark} \\

        \textbf{DVFO (Ours)} & \textcolor{teal}{\checkmark} & \textcolor{teal}{\checkmark}  & \textcolor{teal}{\checkmark}  & \textcolor{teal}{\checkmark} \\ \hline
    \end{tabular}
\end{table}

In order to achieve energy-efficient DNN inference, in this paper, we propose DVFO, a DVFS enabled learning-based collaborative inference framework that automatically co-optimizes the CPU, GPU and memory frequencies of edge devices, as well as the DNN feature maps to be offloaded to cloud servers.
We need to deal with the following issues to design and implement such a framework. 
Firstly, edge-cloud collaborative inference has dynamic network conditions and intense real-time requirements. Deep reinforcement learning (DRL) is effective in dealing with high-dimensional decision and optimization problems, but existing methods applied to edge-cloud collaboration are inefficient to deal with the real-world dynamic environments, e.g., online policy inference cannot catch dynamic environment changes~\cite{andrychowicz2020learning}. 
Thus, we utilize a concurrency mechanism, called \emph{thinking-while-moving}~\cite{xiao2019thinking}, to accelerate policy inference for agents in DRL, as we discuss in details in Section~\ref{DRL}.
Secondly, the feature maps to be offloaded to cloud servers would have a network bottleneck, which can dramatically increase transmission latency and energy consumption.
We leverage a \emph{spatial-channel attention} mechanism~\cite{woo2018cbam} to guide feature maps offloading~\cite{huang2022real}, so that the end-to-end latency can be significantly reduced without sacrificing DNN inference accuracy.
After solving these issues, we perform experiments and compare DVFO with state-of-the-art methods on CIFAR-100~\cite{krizhevsky2009learning} and ImageNet-2012~\cite{krizhevsky2017imagenet} datasets. 
Extensive evaluations show that DVFO can efficiently balance energy consumption and end-to-end latency by automatically co-optimizing the hardware resources of edge devices and the feature maps to be offloaded to cloud servers.

In summary, we make the following contributions:
\begin{itemize}
\item
We propose DVFO, a novel DVFS enabled edge-cloud collaborative DNN inference framework that automatically co-optimizes the hardware frequencies of edge devices, and the proportion of the feature maps to be offloaded to cloud servers.  
\item 
We apply the \emph{thinking-while-moving} concurrent control mechanism in learning-based optimization, and we design an importance-based feature maps offloading scheme to alleviate edge-cloud network bottlenecks by leveraging a \emph{spatial-channel attention} mechanism.
\item 
Extensive evaluations on three heterogeneous edge devices with two datasets show that DVFO reduces energy consumption by up to 33\% on average for various DNN models, compared to state-of-the-art schemes. DVFO also achieves 28.6\%$\sim$59.1\% end-to-end latency reduction, without scarifying accuracy.
\end{itemize}

The rest of the paper is organized as follows: 
Section~\ref{motivation} highlights our research motivations.
Section~\ref{preliminaries} briefly describes deep reinforcement learning we used.
Section \ref{overview} describes system overview and problem formulation. 
Section~\ref{design} illustrates our framework design in detail. 
Section~\ref{evaluation} reports experimental results. 
Section~\ref{Related} presents related work. 
Section~\ref{conclusion} concludes our work.

\section{Motivation} \label{motivation}
Although DNN models can provide state-of-the-art performance for many IoT applications, it comes at the cost of intensive complexity and prohibitive energy consumption. 
Therefore, it is critical to be able to efficiently execute DNN on resource-constrained edge devices. 
In this section, we discuss the experiments and observations that motivate us to develop an efficient DVFS enabled learning-based edge-cloud collaborative inference framework. 
\begin{figure}
\vspace{0pt}
\setlength{\abovecaptionskip}{0pt}
\setlength{\belowcaptionskip}{0pt}
\centering
\subfigure[Jetson Nano with EfficientNet-B0]{
\begin{minipage}[b]{0.48\linewidth}
\includegraphics[width=1\linewidth]{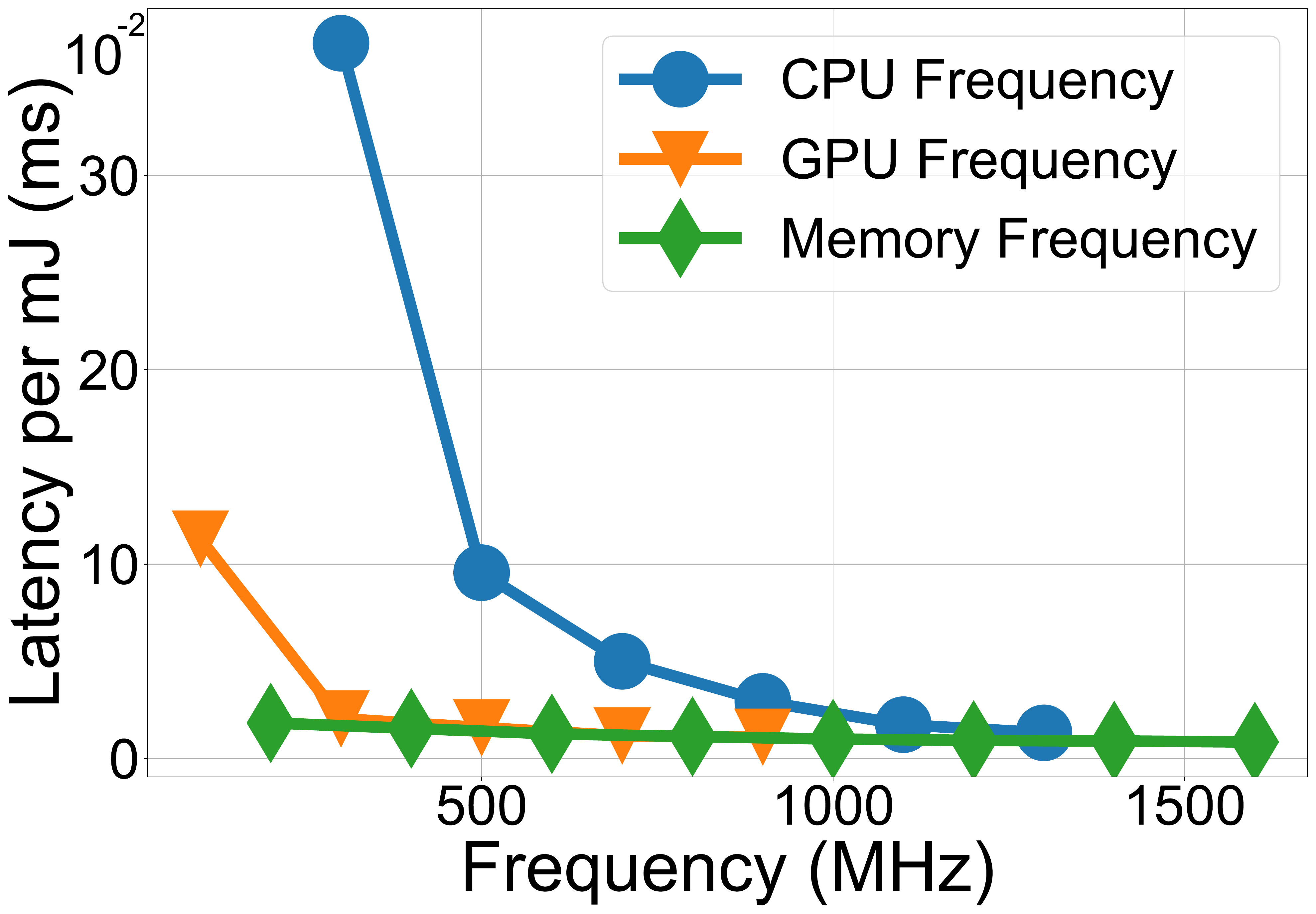}
\end{minipage}}
\subfigure[Xavier NX with EfficientNet-B0]{
\begin{minipage}[b]{0.48\linewidth}
\includegraphics[width=1\linewidth]{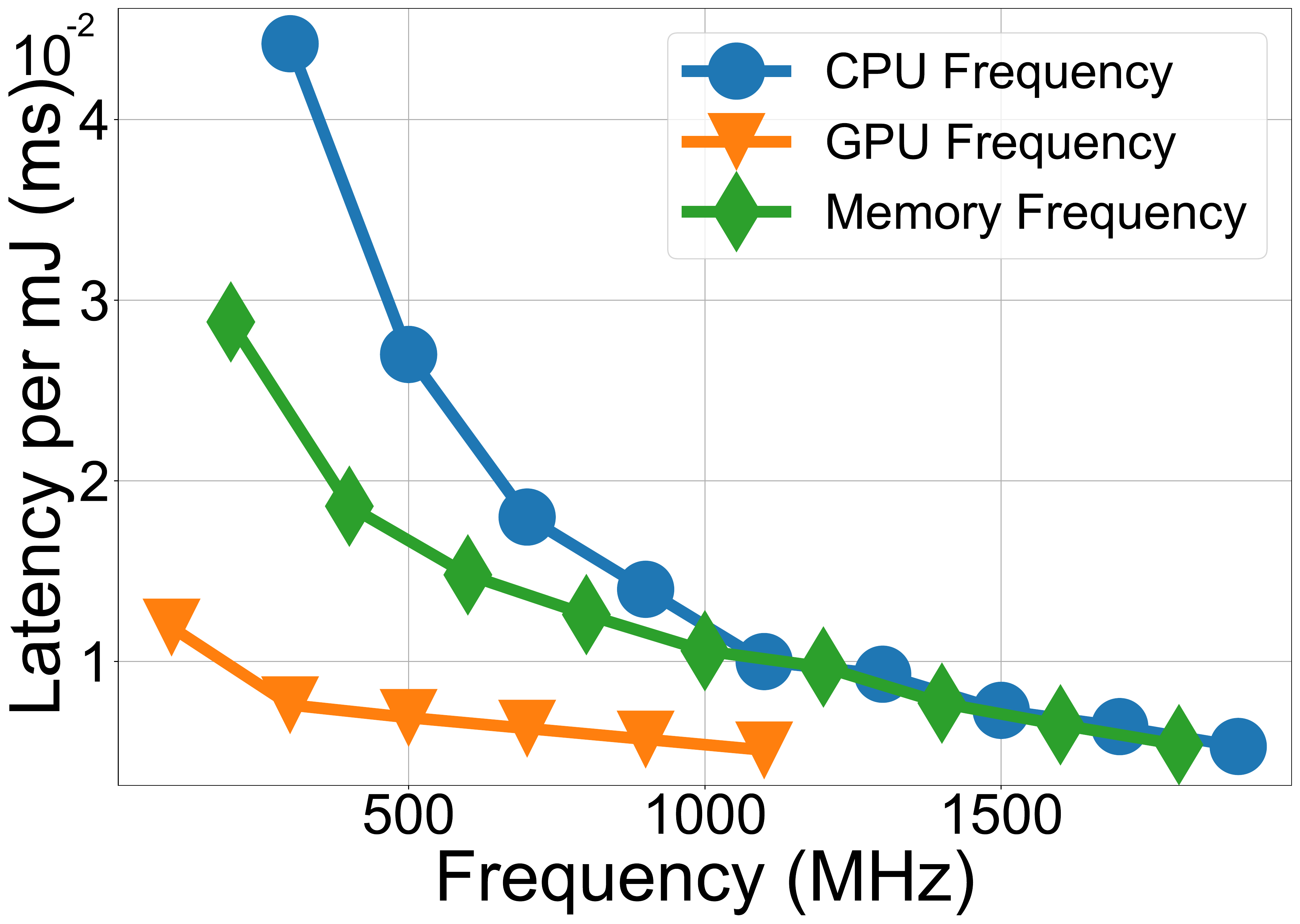}
\end{minipage}}
\subfigure[Jetson Nano with ViT-B16]{
\begin{minipage}[b]{0.48\linewidth}
\includegraphics[width=1\linewidth]{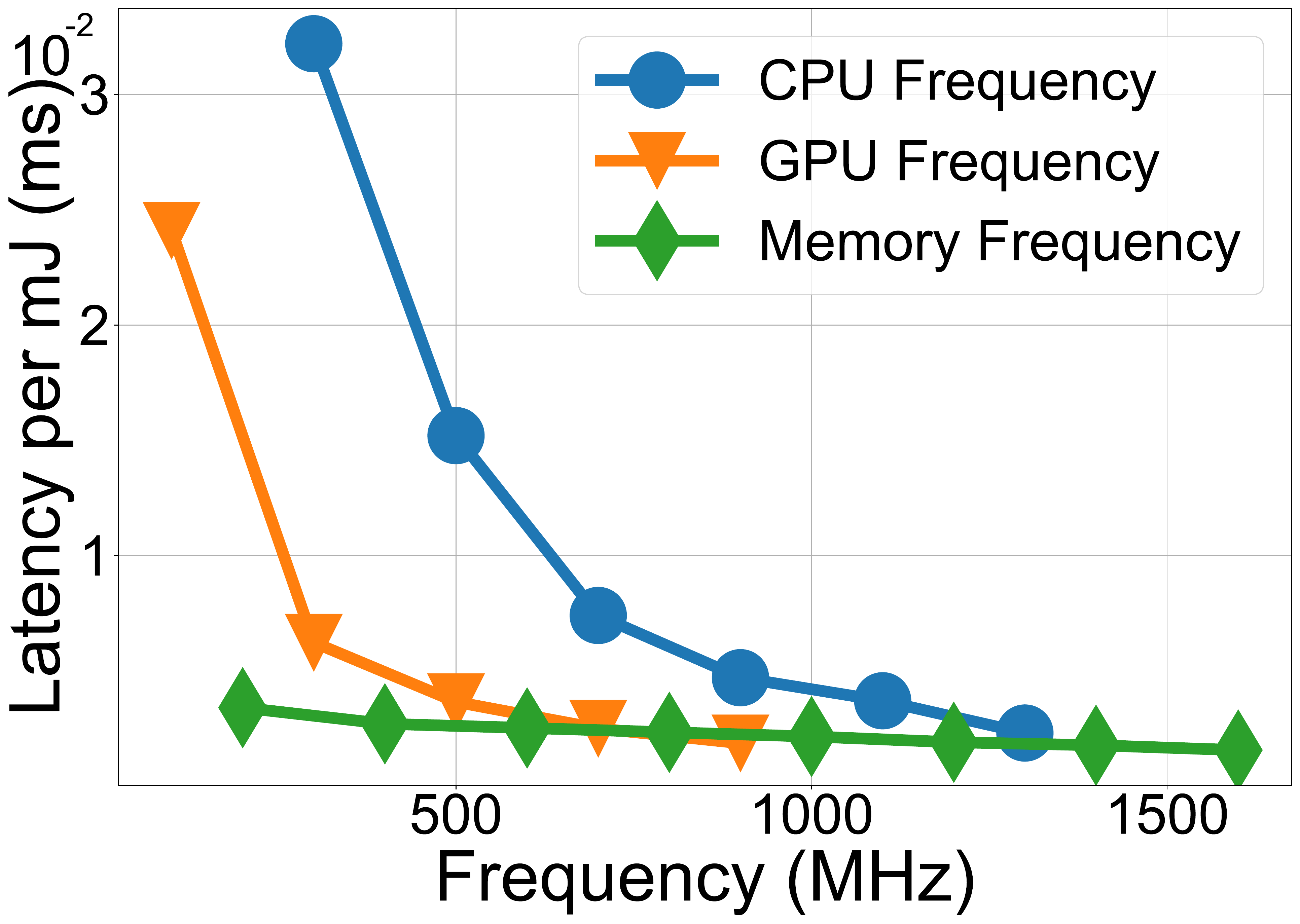}
\end{minipage}}
\subfigure[Xavier NX with ViT-B16]{
\begin{minipage}[b]{0.48\linewidth}
\includegraphics[width=1\linewidth]{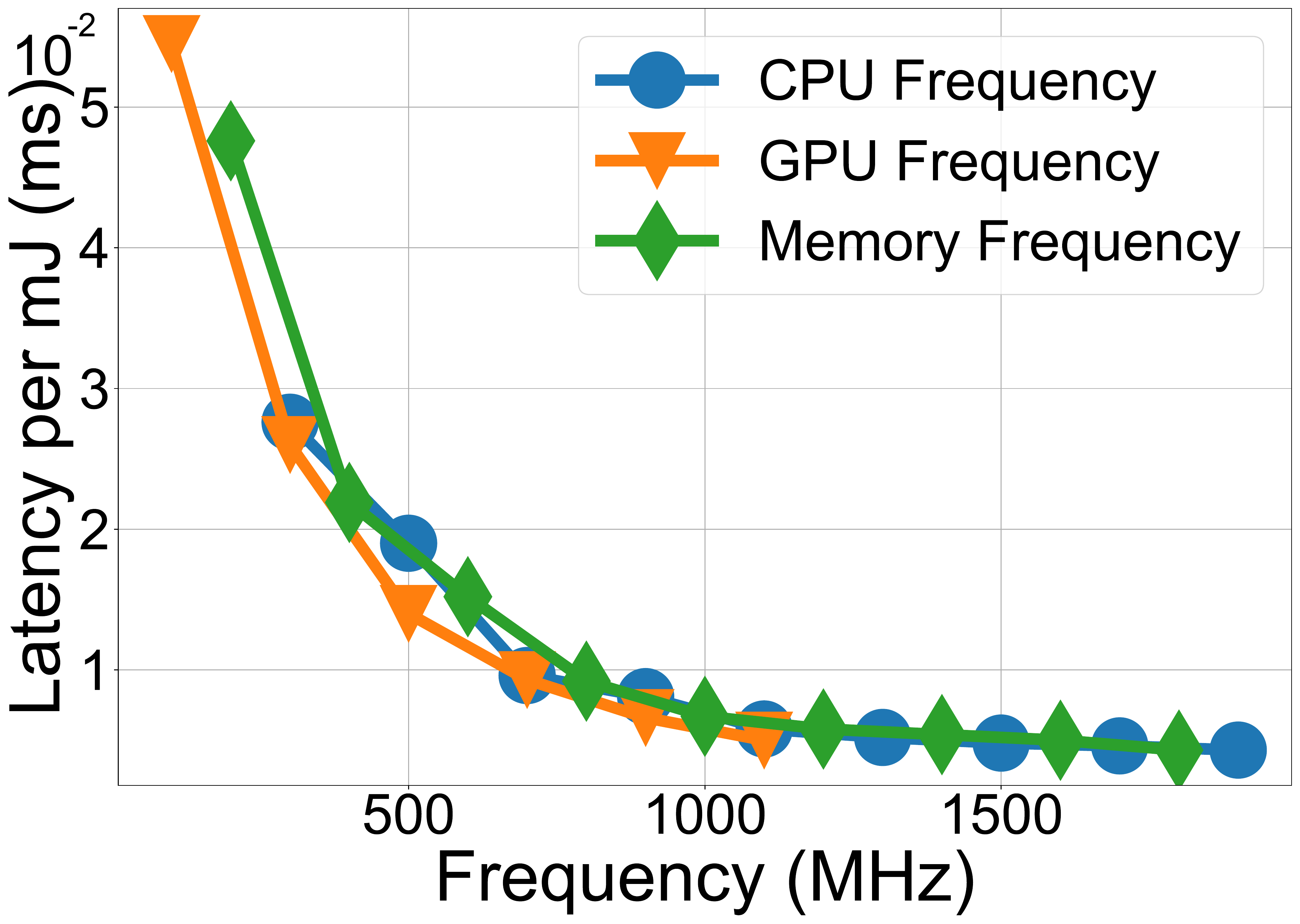}
\end{minipage}}
\caption{The inference performance (i.e., latency per mJ) of three heterogeneous edge devices with different CPU, GPU and memory frequencies for EfficientNet-B0~\cite{tan2019efficientnet} and Visual Transformer (ViT-B16)~\cite{dosovitskiy2020image} DNN models under CIFAR-100~\cite{krizhevsky2009learning} dataset. We set the batch size to 1.}
\label{fig:observation}
\end{figure}

As mentioned in Section~\ref{sec:introduction}, we perform experiments with four widely-deployed DNN models (i.e., ResNet-18~\cite{he2016deep}, MobileNet-v2~\cite{sandler2018mobilenetv2}, EfficientNet-B0~\cite{tan2019efficientnet} and ViT-B16~\cite{dosovitskiy2020image}), and observe that GPU consumes more energy than CPU during the DNN inference phase on edge devices. 
To better understand the impact of CPU, GPU and memory frequencies of edge devices on the end-to-end latency and energy consumption, we further conduct the following experiments and analysis in Fig.~\ref{fig:observation}. 
As you can see, we execute memory-intensive DNN model (e.g., EfficientNet-B0~\cite{tan2019efficientnet}) and compute-intensive (e.g., ViT-B16~\cite{dosovitskiy2020image}) DNN model~\cite{williams2009roofline} on an NVIDIA Jetson Nano and NVIDIA Xavier NX edge platform, respectively.

Note that prior work only considers end-to-end latency or energy consumption as a single metric, which cannot directly reveal the trade-off between inference performance and energy requirements. 
We report the inference performance \emph{latency per mJ}, a metric by dividing end-to-end latency by energy consumption.
As shown in Fig.~\ref{fig:observation}, we measure the inference performance of two heterogeneous edge devices with two aforementioned DNN models under CIFAR-100~\cite{krizhevsky2009learning} dataset using different CPU, GPU and memory frequencies.
We have the following key observations from our experiments and analysis:
\begin{itemize}
\item
\textbf{High frequency does not mean high inference performance.} 
Intuitively, the higher frequency is, the larger amounts of energy the system consumes. 
However, increasing frequency does not improve inference performance (i.e., latency per mJ). 
Take EfficientNet-B0~\cite{tan2019efficientnet} as an example, the energy consumption with the maximum frequency doubled after 500MHz, but the end-to-end latency is not significantly reduced, which means that the inference performance tends to saturate.
Similar phenomenon can be observed for Vision Transformer (ViT-B16)~\cite{dosovitskiy2020image}.
Therefore, a learning approach is needed to automatically find the appropriate hardware frequencies to achieve optimal inference performance.
\item
\textbf{DNN models with different operation intensities exhibit significant end-to-end latency and energy differences on heterogeneous edge devices.}
Take for example the NVIDIA Xavier NX edge platform, which has abundant compute resources.
According to operational density in the roofline model~\cite{williams2009roofline}, we can conclude from the Fig.~\ref{fig:observation}(b) that EfficientNet-B0~\cite{tan2019efficientnet} is a memory-intensive DNN, because the performance bottleneck depends on the CPU and memory frequencies. 
The ViT-B16~\cite{dosovitskiy2020image} with higher complexity in Fig.~\ref{fig:observation}(d) is a compute-intensive DNN model, where GPU frequency dominates performance. 
However, these two DNN models are both compute-intensive on Jetson Nano, which has limited compute resources compared with Xavier NX. 
Thus, it illustrates that the same DNN model exhibit high heterogeneity for edge devices with different computing resources, and DVFS alone cannot further improve inference performance.
Therefore, we highlight that identifying the behavior of various DNN models under heterogeneous devices can further improve the performance of DNN inference.
In addition, we propose to take advantage of the abundant resources on cloud servers to allocate corresponding compute resources to DNN models. 
\end{itemize}

Based on our observations, we highlight two schemes that can deal with the problems, and achieve the trade-off between energy consumption and end-to-end latency in energy-efficient DNN inference: 
(1) dynamic voltage and frequency scaling (DVFS) and 
(2) edge-cloud collaborative inference. Note that the above two schemes are orthogonal. DVFS adjusts hardware frequency to increase end-to-end latency while reducing energy consumption, while edge-cloud collaborative inference can effectively reduce end-to-end latency and further reduce energy consumption. 
To summarize, the observations and analysis motivate us to automatically co-optimize these two aspects for better energy saving and less end-to-end latency.

\section{Preliminaries}\label{preliminaries}
Deep reinforcement learning (DRL) combines deep learning and reinforcement learning, where reinforcement learning is used to define problems and optimize objectives, and deep learning is used to solve the modeling of policies and value functions in reinforcement learning.
In general, DRL uses the back-propagation algorithm to optimize the objective function, which is suitable for solving complex high-dimensional sequential decision problems and achieves impressive performance on many tasks. 
The agent in DRL is used to perceive the environment and make decisions, which performs a task by interacting with the external environment.
Meanwhile, the environment changes its state by responding to the actions selected by the agent, and feeds back corresponding reward signals to the agent.

\begin{figure}[htbp]
\large
\centerline{\includegraphics[width=\linewidth]{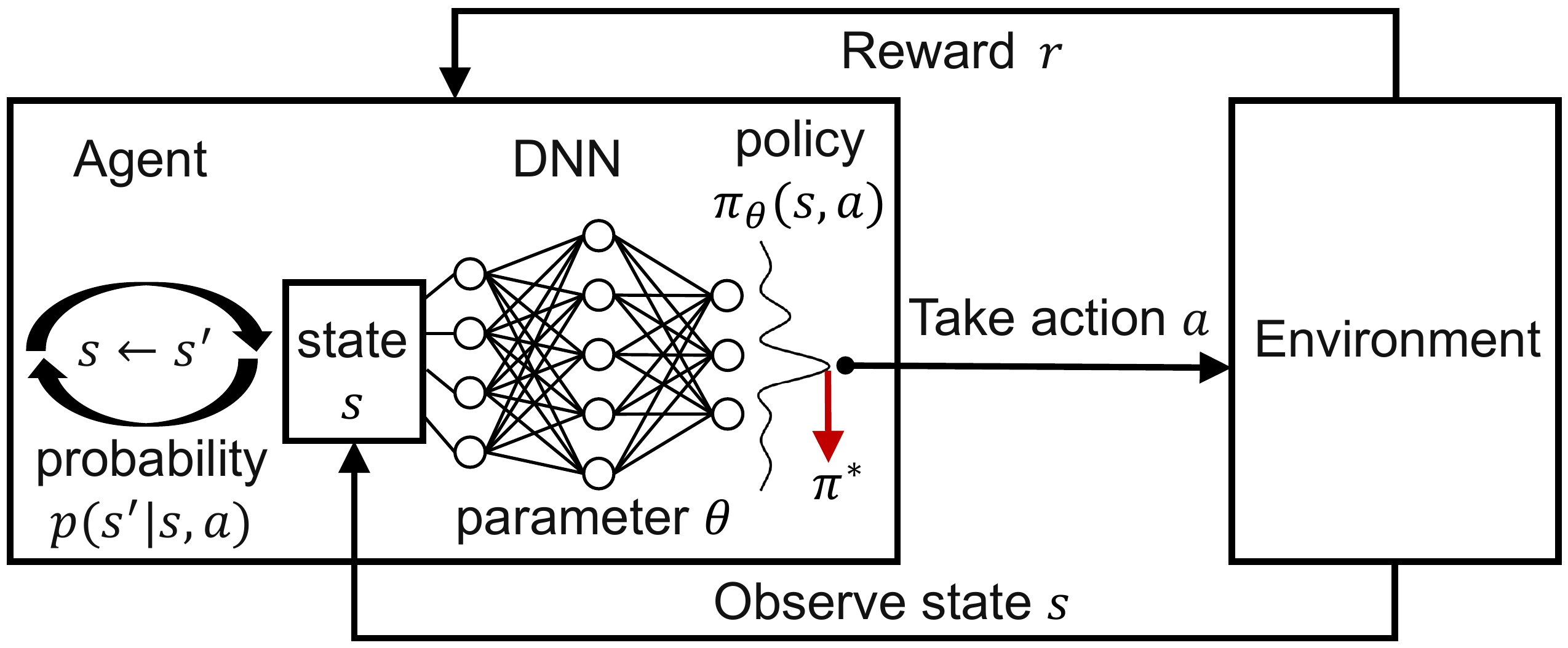}}
\caption{Overview of deep reinforcement learning system.}
\label{fig:DQN}
\end{figure}

As shown in Fig.~\ref{fig:DQN}, most DRL algorithms take the optimization problem as a markov decision process (MDP), which can be described by a tuple: $(\mathcal{S}, \mathcal{A}, \pi, r, p)$, 
where $\mathcal{S}$ is the state space containing all states $s (s\in\mathcal{S})$; 
$\mathcal{A}$ is the action space containing all actions $a (a\in\mathcal{A})$; 
$\pi$ is the probability distribution function that determines the next action $a$ according to the state $s$, satisfying $\sum_{a \in \mathcal{A}} \pi(a|s)=1$; 
$r$ is a scalar function, which means that after the agent makes an action $a$ according to the current state $s$, the environment feeds back a reward signal to the agent.
Note that $r$ is related to the state $s^{\prime}$ at the next moment due to hysteresis;
$p$ is the state transition probability, which means that after the agent makes an action $a$ according to the current state $s$, the probability that the environment changes to the state $s^{\prime}$ at the next moment, also satisfying $\sum_{s^{\prime} \in \mathcal{S}} p(s^{\prime}|s,a)=1$.

The goal of the DRL algorithm is to find an optimal policy $\pi^{*}$ to maximize the following expected return:
\begin{equation}
\begin{split}
\pi^{*}=\operatorname{argmax}_{\theta}\mathbb{E}_{\tau \sim p(\tau)}\left[\sum_{t=0}^{T-1} \gamma^{t-1} r_t\right],
\label{max-reward}
\end{split}
\end{equation}
where $\tau=s_0,a_0,r_0,s_1,a_1,r_1,\cdots, s_{T-1},a_{T-1},r_{T-1}$ is a trajectory that represents an interaction process between the agent and the environment.
$\theta$ is the parameter of policy network, and $\gamma \in [0,1]$ is a discount factor.
We can obtain the optimal policy $\pi^{*}=\operatorname{argmax}_{a}Q^{*}(s,a)$ by value iteration via the following the Bellman optimal equation of state-action value function (Q-function):
\begin{equation}
\begin{split}
Q^{*}(s,a)=\mathbb{E}_{\tau \sim p(\tau)}[r(s_t,a_t)+\gamma\operatorname{max}_{a_{t+1}}Q^{*}(s_{t+1},a_{t+1})]
\label{Q-value}
\end{split}
\end{equation}

In Section~\ref{DRL}, we describe the DQN-based DRL algorithm in detail.

\begin{figure*}[htbp]
\large
\centerline{\includegraphics[width=0.7\linewidth]{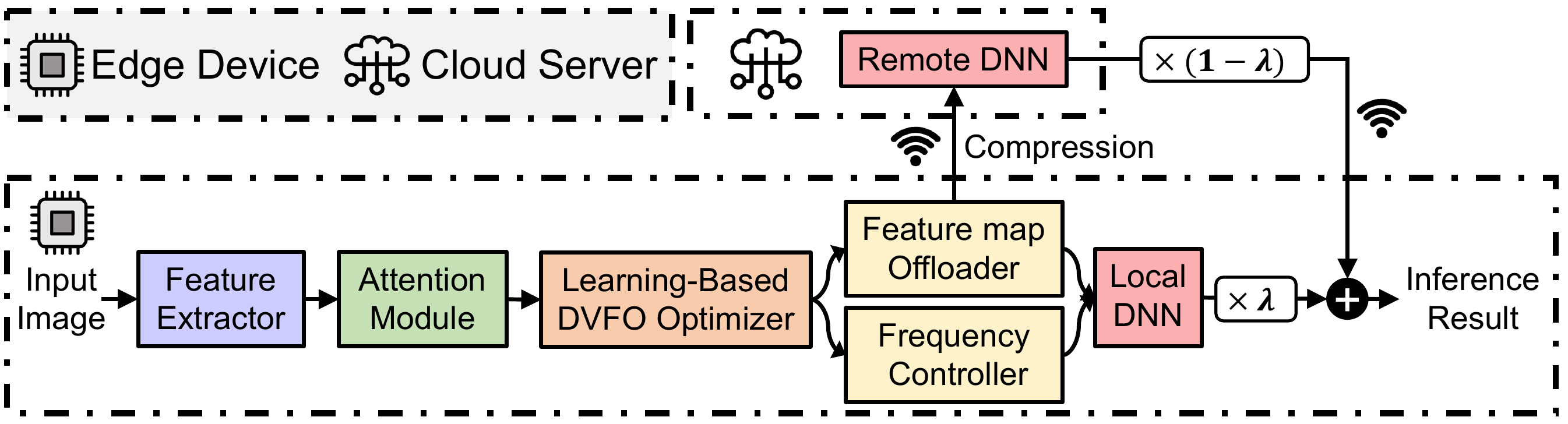}}
\caption{Overview of the architecture of DVFO framework.}
\label{fig:system}
\end{figure*}

\section{System Overview and Problem Statement} \label{overview}
\subsection{System Overview}
Fig.~\ref{fig:system} shows an overview of our DVFO framework. The framework incorporates local DNN inference on edge devices and remote DNN inference on cloud servers. 
During DNN inference, users submit DNN inference tasks to DVFO, along with user-defined parameters to adjust the trade-off between energy consumption and end-to-end latency (i.e., the weight parameter $\eta$ in Eq.~(\ref{metric})), and the workflow starts as follows. 
\ding{182} DVFO utilizes a feature extractor on edge devices to extract high-dimensional features of the input data and obtain DNN feature maps. 
The feature extractor is implemented based on a lightweight neural network with negligible overhead.
\ding{183} To alleviate network bottlenecks of the feature maps to be offloaded to cloud servers, DVFO utilizes \emph{spatial-channel attention} module to evaluate the importance of feature maps, in order to guide the feature maps offloading. The attention module details are in Section~\ref{attention}.
\ding{184} The DRL-based DVFO module (i.e., DVFO optimizer) learns the optimal hardware frequency vector and the proportion parameter of the feature maps to be offloaded to cloud servers for each task based on historical data, current bandwidth, and user configuration (see Section~\ref{DRL} for more details).
\ding{185} Based on the optimal hardware frequencies and the feature maps to be offloaded to cloud servers learned by DVFO optimizer, DVFO retains the top-k features with primary-importance for local DNN inference, and then combines the remote DNN with other compressed secondary-importance features via weighted summation (the summation weight $\lambda \in (0,1)$ can also be user-defined), to produce the final prediction result on edge devices locally. 
Compared to adding additional neural network (NN) layers for fusion, such a point-to-point weighted summation method is much more lightweight and has low computation overhead on edge~\cite{huang2022real}. 

\subsection{Problem Statement}\label{Problem}
Opportunities to reduce the energy consumption of DNN inference come at the cost of increased end-to-end latency. 
When optimized for energy consumption, DNN end-to-end latency (i.e., time-to-inference, or $\operatorname{TTI}$) may be impaired.
Here we define the energy consumption of DNN inference as its energy-to-inference (ETI):
\begin{equation}
\begin{split}
\operatorname{ETI}(\mathbf{f},\xi) = \operatorname{TTI}(\mathbf{f},\xi) \times \operatorname{AvgPower}(\mathbf{f},\xi),
\label{energy}
\end{split}
\end{equation}
where $\mathbf{f}$ and $\xi$ are the hardware frequency vector of device, and the proportion of the feature maps to be offloaded to cloud servers, respectively, and $\operatorname{AvgPower}$ is the average power consumption during inference with configuration ($\mathbf{f},\xi$). 
Different from prior work~\cite{panda2022energy} that only considers the CPU frequency $f^C$, we also incorporate GPU and memory frequencies of edge devices, denoted as $f^G$, $f^M$, respectively, that is, $\mathbf{f}=(f^C, f^G, f^M)$. 

\textbf{Cost metric}: It is important to define a cost metric in designing DVFO, so that users can adjust the trade-off between energy consumption and end-to-end latency based on the application requirements and their preferences.
Thus we propose the following cost metric:
\begin{equation}
\begin{split}
C(\mathbf{f},\xi;\eta) = \eta \cdot \operatorname{ETI}(\mathbf{f},\xi) + (1-\eta) \cdot \operatorname{MaxPower} \cdot \operatorname{TTI}(\mathbf{f}, \xi),
\label{metric}
\end{split}
\end{equation}
where $\operatorname{MaxPower}$ is the maximum power limit supported by device, a constant introduced to unify the units of measure in the cost metric~\cite{you2023zeus}, and $\eta\ \in[0,1]$ is a weight parameter that users define to adjust the balance between energy consumption and end-to-end latency. 
In particular, when $\eta=0$, we are only optimizing energy consumption $\operatorname{ETI}$, whereas when $\eta=1$, only end-to-end latency $\operatorname{TTI}$ is optimized. 
A more detailed sensitivity analysis of the parameter $\eta$ can be found in Section~\ref{evaluation}.

\textbf{End-to-end latency model}: For a set of DNN inference tasks $\mathcal{X}={(x_1, x_2, ..., x_N)}$ consisting of $N$ independent and non-preemptive tasks $x_i$, $i=1,\cdots,N$.
We show the optimization problem in terms of end-to-end latency and energy consumption. 
First, for end-to-end latency $\operatorname{TTI}_i^{total}$, it incorporates 
1) the computing time on edge for the $i$-th task $\operatorname{TTI}_i^{local}$, 
2) the compression (quantization) time of the feature map to be offloaded to cloud servers on edge $\operatorname{TTI}_i^{comp}$,
3) the transmission time of the offloaded feature maps to cloud $\operatorname{TTI}_i^{off}$, and 
4) the computing time on cloud $\operatorname{TTI}_i^{cloud}$.
Note that we ignore the fusion time on edge devices and the decompression time on cloud servers, benefit from the lightweight weighted summation-based fusion method on edge devices in Section~\ref{combine} and the abundant computing power of the cloud servers, respectively.
Specifically, the computing time on edge $\operatorname{TTI}_i^{local}$ depends on two factors: the size of feature maps without offloading $m_i^{local}$, and the hardware frequency of edge devices $(f_{local}^C, f_{local}^G, f_{local}^M)$, which can be defined as:
\begin{equation}
\begin{split}
\operatorname{TTI}_i^{local} = \frac {m_i^{local}} {(f_{local}^C, f_{local}^G, f_{local}^M)},
\label{c_local}
\end{split}
\end{equation}

Likewise, the computing time on cloud $\operatorname{TTI}_i^{cloud}$ depends on the size of the feature maps to be offloaded to cloud servers $m_i^{cloud}$, and the hardware frequency of cloud servers $(f_{cloud}^C, f_{cloud}^G, f_{cloud}^M)$
\begin{equation}
\begin{split}
\operatorname{TTI}_i^{cloud} = \frac {m_i^{cloud}} {(f_{cloud}^C, f_{cloud}^G, f_{cloud}^M)},
\label{c_cloud}
\end{split}
\end{equation}

The compression time on edge $\operatorname{TTI}_i^{comp}$ depends on the size of the feature maps to be offloaded to cloud servers $m_i^{cloud}$. 
In this work, we use quantization aware training (QAT) in Section~\ref{implementation} to effectively compress the offloaded feature maps with low-bit quantization (i.e., converted from float-32 model to int-8 model).
The compression time on edge $\operatorname{TTI}_i^{comp}$ defined as
\begin{equation}
\begin{split}
\operatorname{TTI}_i^{comp} = \operatorname{QAT}(m_i^{cloud}),
\label{c_comp}
\end{split}
\end{equation}

The transmission time $\operatorname{TTI}_i^{off}$ is affected by the size of the feature maps to be offloaded to cloud servers $m_i^{cloud}$ and the communication bandwidth $\mathcal{B}$, that is
\begin{equation}
\begin{split}
\operatorname{TTI}_i^{off} = \frac {m_i^{cloud}} {\mathcal{B}},
\label{c_off}
\end{split}
\end{equation}

Note that the size of the feature maps to be offloaded to cloud servers $m_i^{cloud}$ is determined by the proportion parameter $\xi$ in Eq.~(\ref{metric}).

Therefore, the end-to-end latency $\operatorname{TTI}_i^{total}$ can be formulated as follows
\begin{equation}
\begin{split}
\operatorname{TTI}_i^{total} = \operatorname{TTI}_i^{local}+\operatorname{TTI}_i^{comp}+\operatorname{TTI}_i^{off}+\operatorname{TTI}_i^{cloud}
\label{c_total}
\end{split}
\end{equation}

\textbf{Energy consumption model}: For energy consumption, the overall energy consumption $\operatorname{ETI}_i^{total}$ of edge devices for a particular task $x_i$ consists of the energy consumption for computing $\operatorname{ETI}_i^c$ and the energy consumption for offloading $\operatorname{ETI}_i^o$, that is 
\begin{equation}
\begin{split}
\operatorname{ETI}_i^{total}=\operatorname{ETI}_i^c+\operatorname{ETI}_i^o
\label{e_total}
\end{split}
\end{equation} 

To be more specific, the energy consumption for computing $\operatorname{ETI}_i^c$ of $i$-th task $x_i$ depends on the edge computing time $\operatorname{TTI}_i^{local}$ and the computing power of edge devices $p_i^c$, which can be defined as
\begin{equation}
\begin{split}
\operatorname{ETI}_i^c= \operatorname{TTI}_i^c \cdot p_i^c, 
\label{e_compute}
\end{split}
\end{equation} 
where $p_i^c$ is proportional to the square of the voltage $V^2$ and the frequency $\mathbf{f}$, i.e., $p_i^c \propto V^2 \cdot \mathbf{f}_{i}$. 

The energy consumption of offloading $\operatorname{ETI}_i^o$ for $x_i$ is affected by the communication bandwidth $\mathcal{B}$ of the network between edge devices and cloud servers, the proportion of the feature maps to be offloaded to cloud servers $m_i^{cloud}$, and the offloading power of edge devices $p_i^o$, that is 
\begin{equation}
\begin{split}
\operatorname{ETI}_i^o= \frac {m_i^{cloud}\cdot p_i^o} {\mathcal{B}}.
\label{e_offload}
\end{split}
\end{equation}

Similarly, $p_i^o \propto V^2 \cdot \mathbf{f}_{i}$.

The objective of of DVFO is to minimize the cost in Eq.~(\ref{metric}) by automatically exploring the feasible set of edge hardware frequency vector $(f^C, f^G, f^M)$ and the offloading proportion parameter $\xi$, while keeping minimal frequencies $\mathbf{f}_{min}$ at which the system operations while not exceeding the maximum frequency $\mathbf{f}_{max}$. 
Put formally in terms of the cost function defined by Eq.~(\ref{metric}), our objective becomes
\begin{equation}
\begin{split}
\min _{\mathbf{f},\xi}. & C(\mathbf{f},\xi;\eta) \\
s.t.\; \mathbf{f}_{min}  & \leq (f^C, f^G, f^M) \leq \mathbf{f}_{max} \\
     0 & \leq \xi \leq 1
\label{metric1}
\end{split}
\end{equation}

For each task, DVFO can automatically co-optimize CPU, GPU and memory frequencies, as well as the proportion of the feature maps to be offloaded to cloud servers. 
Note that we assume cloud servers have enough compute resources to guarantee the real-time performance of remote inference. 
We also assume that edge devices can be put into idle mode after the inference and offloading operations to save energy.

Table~\ref{symbol} provides the notation and corresponding descriptions used in this paper.
\begin{table}
\setlength{\abovecaptionskip}{0pt}
\setlength{\belowcaptionskip}{0pt}
    \caption{Notation and Description}
    \label{symbol}
    \centering
    \begin{tabular}{l|l} \hline
    	Notation & Description \\ \hline
        $\mathcal{X}$ & the whole task set \\
        $x_i$ & the $i$-th non-preemptive task \\
        $\operatorname{TTI}$ & the time-to-inference  \\
        $\operatorname{ETI}$ & the energy-to-inference  \\
    	$C$ & the cost metric  \\
        $f^C$ & the CPU frequencies of edge devices \\
        $f^G$ & the GPU frequencies of edge devices \\
        $f^M$ & the memory frequencies of edge devices \\
    	$\xi$ & the proportion of the feature maps to be offloaded \\ 
        $\eta$ & the weight parameter \\
        $m_i^{local}$ & the size of feature maps without offloading \\
        $m_i^{cloud}$ & the size of feature maps with offloading \\
        $\mathcal{B}$ & the communication bandwidth \\
        $V^2$ & the voltage of edge devices \\
        $p_i^c$ & the computing power of edge devices \\
        $p_i^o$ & the offloading power of edge devices \\
        $\lambda$ & the summation weight \\ \hline
    \end{tabular}
    \vspace{-0.1in}
\end{table}

\section{System Design} \label{design}
\subsection{Learning-based DVFO}\label{DRL}
In this section, we discuss how DVFO determines the hardware frequency vector $\mathbf{f}$ and the proportion of feature maps to be offloaded $\xi$ for each task, as shown in Eq.~(\ref{metric1}).
Here we formulate the optimization problem as a markov decision process (MDP), and utilize deep reinforcement learning (DRL) to automatically determine the optimal configuration.
We choose DRL since it can efficiently deal with policy decision problems in complex high-dimensional spaces.

More specific, we transform the optimization objective in Eq.~(\ref{metric1}) into a reward in DRL. 
The agent in DRL has three components, namely state, action and reward, which are defined as follows:
\begin{itemize}
\item 
\textbf{\emph{State Space}}: At each time step $t$, the agent in DRL will construct a state space $\mathcal{S}$. We define the weight parameter $\eta$ specified by the user, the adjustable summation weight $\lambda$, the importance distribution of features $\mathbf{x}\sim\mathbf{p}(\mathbf{a})$, and the current network bandwidth $\mathcal{B}$ as state. 
The above measures constitute the state space $\mathcal{S}$, denoted as $\mathcal{S}=\{\lambda, \eta, \mathbf{x}\sim\mathbf{p}(\mathbf{a}), \mathcal{B}\}$. 
\item
\textbf{\emph{Action Space}}: We set the frequency vector $\mathbf{f}_i$ and the offloading proportion parameter $\xi_i$ for $x_i$ as actions. 
Therefore, the action space can be expressed as $\mathcal{A}=\{\mathbf{f}_i,\xi_i\}$, where $\mathbf{f}_i=(f_i^C, f_i^G, f_i^M)$ represents the CPU, GPU and memory frequencies for a particular task $x_i$. 
For example $(1500, 900, 1200, 0.3)$ means that 30\% of feature maps are executed locally, and the remaining of the feature maps are offloaded to the remote, when the CPU, GPU and memory frequencies are set to 1500MHz, 900MHz and 1200MHz, respectively. 
To reduce the complexity of the action space for faster convergence, we set both the frequency and the proportion of feature maps to be offloaded to discrete values. 
Specifically, we evenly sample 100 frequency levels between the minimum frequency that satisfy the system operation and the maximum frequency for the CPU, GPU and memory frequencies, respectively.
\item 
\textbf{\emph{Reward}}: 
Since we need to minimize the cost for each task in Eq.~(\ref{metric1}) with DVFS and edge-cloud collaborative inference by trading off energy consumption and end-to-end latency, the agent in DRL aims to maximize the cumulative expected reward $\mathbb{E}\left[\sum_{t=0}^{T} \gamma^{t} r_{t}\right]$. 
Hence, we transfer the objective of optimizing cost into a reward function, and we define the reward function $r$ as follows:
\begin{equation}
r = -C(\mathbf{f},\xi;\eta).
\label{2}
\end{equation}
\end{itemize}

\begin{figure}[htbp]
\large
\centerline{\includegraphics[width=\linewidth]{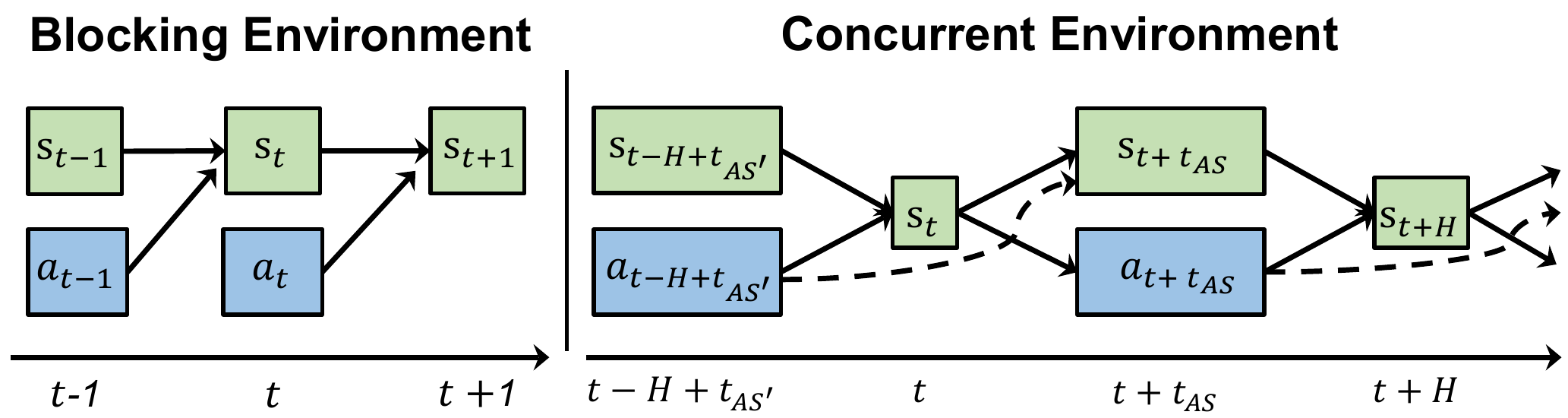}}
\caption{Action trajectories in blocking and concurrent environments.}
\label{fig:think}
\end{figure}

However, as the left part of Fig.~\ref{fig:think} shows, most DRL algorithms assume that the state of the environment is static, in which the agent is making a decision. 
That is, the agent first observes the state and then executes policy inference. However, this blocking approach of sequential execution is not suitable for real-world dynamic real-time environments. 
Because the state has "slipped" after the agent observes the state of the environment and executes an action, i.e., the previous state transitions to a new unobserved state. 
This environment is regarded as a concurrent environment in~\cite{xiao2019thinking}. 
In particular, in the edge-cloud collaboration environment with strict time constraints, we need to use DRL to adjust the frequency of edge devices and the proportion of feature maps to be offloaded in real-time, according to the importance of features and network bandwidth. 
Therefore, it is crucial to reduce the overhead of policy inference in DRL.

To deal with the issues, in this work, we utilize DQN~\cite{mnih2013playing} to learn the optimal CPU, GPU and memory frequencies, as well as the proportion of feature maps to be offloaded for edge devices. 
We use the concurrency control mechanism to reduce the overhead of policy inference in DQN with discrete time based on a \emph{thinking-while-moving} mechanism~\cite{xiao2019thinking}. 
The right part of Fig.~\ref{fig:think} illustrates this concurrent approach. Specifically, the agent observes the state of the environment $s_i$ at time step $t$. 
When it selects an action $a_{t+t_{AS}}$, the previous action $a_{t-H+t_{AS^{\prime}}}$ has slid to a new unobserved state $s_{t+t_{AS}}$, meaning that state capture and policy inference in concurrent environment can be executed concurrently. 
Here $H$ is the duration of the action trajectory from the state $s_t$ to $s_{t+t_{H}}$.

We implement policy inference in concurrent environments by modifying standard DQN.
The concurrent Q-value function of DQN in policy $\pi$ can be reformulated as follows:
\begin{equation}
\begin{split}
& Q^\pi\left(s_t, a_{t-1}, a_t, t, t_{A S}, H\right) = r\left(s_t, a_{t-1}\right) \\
& +\gamma^{\frac{t_{A S}}{H}} Q^\pi\left(s_{t+t_{A S}}, a_t, a_{t+1}, t+t_{A S}, t_{A S^{\prime}}, H-t_{A S}\right).
\label{3}
\end{split}
\end{equation}

Algorithm \ref{alg:alg1} illustrates the optimization process of DVFO in detail. We first initialize the parameters of neural network and replay memory in DRL. 
Then we take $\{\lambda, \eta, \mathbf{x}\sim\mathbf{p}(\mathbf{a}), \mathcal{B}\}$ as the initial state. 
At the start of training, the agent in DRL will select an action randomly. 
In each time step $t$, the agent captures the state $s_t$ in a discrete-time concurrent environment, and chooses an action $a_t$ using a \emph{thinking-while-moving} concurrency mechanism. 
We use the $\epsilon$-greedy strategy to explore the environment. 
Next, we feed the CPU, GPU, and memory frequencies, as well as the proportion of feature maps to be offloaded, selected by the agent to frequency controller and feature maps offloader, respectively. 
Simultaneously, the agent obtains an instant reward $r$, and the state changes from $s_t$ to $s_{t+1}$.
We store the current state, action, reward, and the state of the next time step as a transition in the replay memory. 
At each gradient step, we first sample mini-batch transitions from replay memory randomly. 
Then we use Eq.~(\ref{3}) to calculate the Q-value in the concurrent environment and update the network parameters using gradient descent. 
Finally, we deploy the trained DVFO online to evaluate the performance. Note that the training process is offline.
\begin{algorithm}
\SetKwInOut{Input}{Input}
\SetKwInOut{Output}{Output}
\caption{DVFO Optimization Process}
\label{alg:alg1}
\Input{user preference $\lambda, \eta$; feature maps importance $\mathbf{x}\sim\mathbf{p}(\mathbf{a})$, and current network bandwidth $\mathcal{B}$}
\Output{the optimal settings of hardware frequency $\mathbf{f}_i$ and offloaded proportion $\xi_i$ for each task $x_i$}
Initialize the parameters of network $Q$ and target network $Q^{\prime}$ with $\theta_1$ and $\theta_2$, respectively\;
Initialize an empty replay memory $\mathcal{D} \leftarrow \varnothing$\;
Observe state state $s_0=\{\lambda, \eta, \mathbf{x}\sim\mathbf{p}(\mathbf{a}), \mathcal{B}\}$\;
Initialize action $a_0=\{\mathbf{f}_0,\xi_0\}$ randomly\;
\For{environment step $t\leftarrow 1$ \KwTo $T$}{
    \For{the $i$-th stage $i\leftarrow 1$ \KwTo $N$}{
        Observe state $s_t$ in concurrent environment\;
        Select an action $a_t$ using \emph{thinking-while-moving} with $\epsilon$-greedy\;
        Feed frequency controller and feature maps offloader, respectively\;
        Execute computing and offloading and obtain reward $r$ by Eq.~(\ref{2})\;
        Set $s_t \leftarrow s_{t+1}$\;
        Store transition ${(s_{t}, a_{t}, r(s_{t}, a_{t}), s_{t+1})}$ in $\mathcal{D}$\;
    }
}
\For{each gradient step $g\leftarrow 1$ \KwTo $G$}{
    Sample minibatch of transitions form $\mathcal{D}$\;
    Calculate Q-value using Eq.~(\ref{3})\;
    Update $\theta_1$ of by gradient descent\;
}
\end{algorithm}

\subsection{Spatial-Channel Attention Module}\label{attention}
\begin{figure*}[htbp]
\large
\centerline{\includegraphics[width=0.85\linewidth]{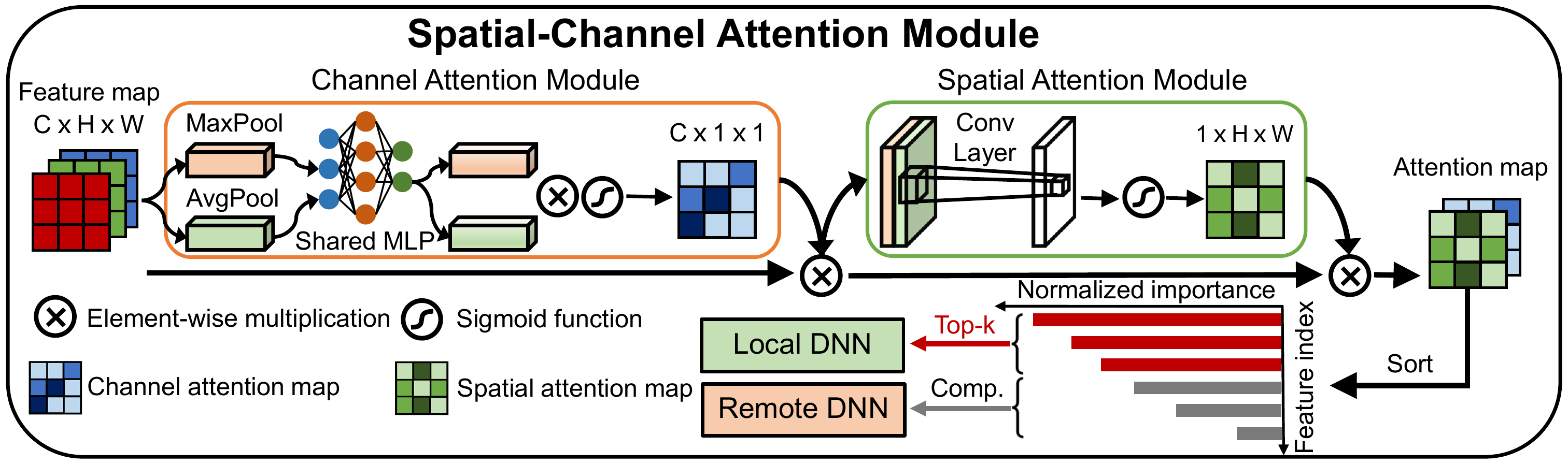}}
\caption{The overview of \emph{spatial-channel attention} module (SCAM). The module has two sequential sub-modules: \emph{channel attention} module and \emph{spatial attention} module. 
The intermediate feature maps are divided into tok-k primary-importance and remaining secondary-importance feature maps by SCAM, and they are executed by local DNN and remote DNN, respectively.}
\label{fig:attention}
\end{figure*}

The effectiveness of offloading in DVFO depends on the skewness~\cite{huang2022real} of the importance distribution of feature maps. 
The higher the skewness, the fewer features dominate DNN inference. 
Therefore, we leverage a \emph{spatial-channel attention} mechanism, namely \emph{spatial-channel attention} module (SCAM) as shown in Fig.~\ref{fig:attention}, to evaluate the feature importance of input data.
Attention is a widely used deep learning technique that allows a network to focus on relevant parts of the input, and suppress irrelevant ones. 
We use it to identify features of primary importance and features of secondary importance in order to guide feature maps offloading.

In this way, we can reduce transmission latency by offloading the compressed secondary-importance features without significantly sacrificing the accuracy of DNN models.

Given a feature maps $\mathbf{F} \in \mathbb{R}^{C \times H \times W}$ extracted by feature extractor as input, SACM sequentially infers a 1D channel attention map $\mathbf{M_c} \in \mathbb{R}^{C \times 1 \times 1}$ and a 2D spatial attention map $\mathbf{M_s} \in \mathbb{R}^{1 \times H \times W}$.
For the arrangement of sequential process, experimental results in~\cite{woo2018cbam} show that channel-first is better than spatial-first.
We next describe the details of each module.

\subsubsection{Channel Attention Module}
In general, since each channel of a feature maps in DNN is considered as a feature detector, the channel attention module in SCAM focuses on "what" is meaningful given an input data.
To fully extract richer channel attention, we aggregate the spatial information of the feature maps using average pooling ($\operatorname{AvgPool}$) and max pooling ($\operatorname{MaxPool}$).
We then feed the generated average-pooled features and max-pooled features into a shared network consisting of multi-layer perceptron ($\operatorname{MLP}$) to obtain channel attention map. 
The channel attention is computed as follows:
\begin{equation}
\begin{aligned}
\mathbf{M}_{\mathbf{c}}(\mathbf{F}) & =\sigma(\operatorname{MLP}(\operatorname{AvgPool}(\mathbf{F}))+\operatorname{MLP}(\operatorname{MaxPool}(\mathbf{F})))
\end{aligned}
\label{attention_1}
\end{equation}
where $\sigma$ denotes the sigmoid function. 

\subsubsection{Spatial Attention Module}
As a complement to \emph{channel attention}, \emph{spatial attention} focuses on "where" is an informative part. We also use average pooling and max pooling along the channel axis to aggregate spatial information of feature maps. 
The generated average pooling features and max pooling features are concatenated and convolved by a 3$\times$3 convolutional layer to generate a spatial attention map.
The spatial attention is computed as follows:
\begin{equation}
\begin{aligned}
\mathbf{M}_{\mathbf{s}}(\mathbf{F}) & =\sigma\left(Conv(3,3)[\operatorname{AvgPool}(\mathbf{F}) ; \operatorname{MaxPool}(\mathbf{F})]\right)
\end{aligned}
\label{attention_2}
\end{equation}
where $Conv(3,3)$ represents a convolution operation with a filter size of 3$\times$3.

\textbf{Arrangement of attention modules.}
Based on the channel attention map and spatial attention map obtained by Eq.~(\ref{attention_1}) and Eq.~(\ref{attention_2}), we can obtain the final attention map $\mathbf{F}^{out}$ by element-wise multiplication.
\begin{equation}
\begin{aligned}
\mathbf{F}^{in} & =\mathbf{M}_{\mathbf{c}}(\mathbf{F}) \otimes \mathbf{F}, \\
\mathbf{F}^{out} & =\mathbf{M}_{\mathbf{s}}\left(\mathbf{F}^{in}\right) \otimes \mathbf{F}^{in}
\end{aligned}
\label{attention_3}
\end{equation}
where $\otimes$ denotes element-wise multiplication, $\mathbf{F}^{in}$ is the intermediate attention map.
We can derive the importance distribution of features $\mathbf{x}\sim\mathbf{p}(\mathbf{a})$ from the normalized weights in final attention map $\mathbf{F}^{out}$, where $\mathbf{x}$ represents the feature maps index, and $\mathbf{a}\in(0,1)$ is the normalized feature importance.

\begin{figure}[htbp]
\large
\centerline{\includegraphics[width=0.9\linewidth]{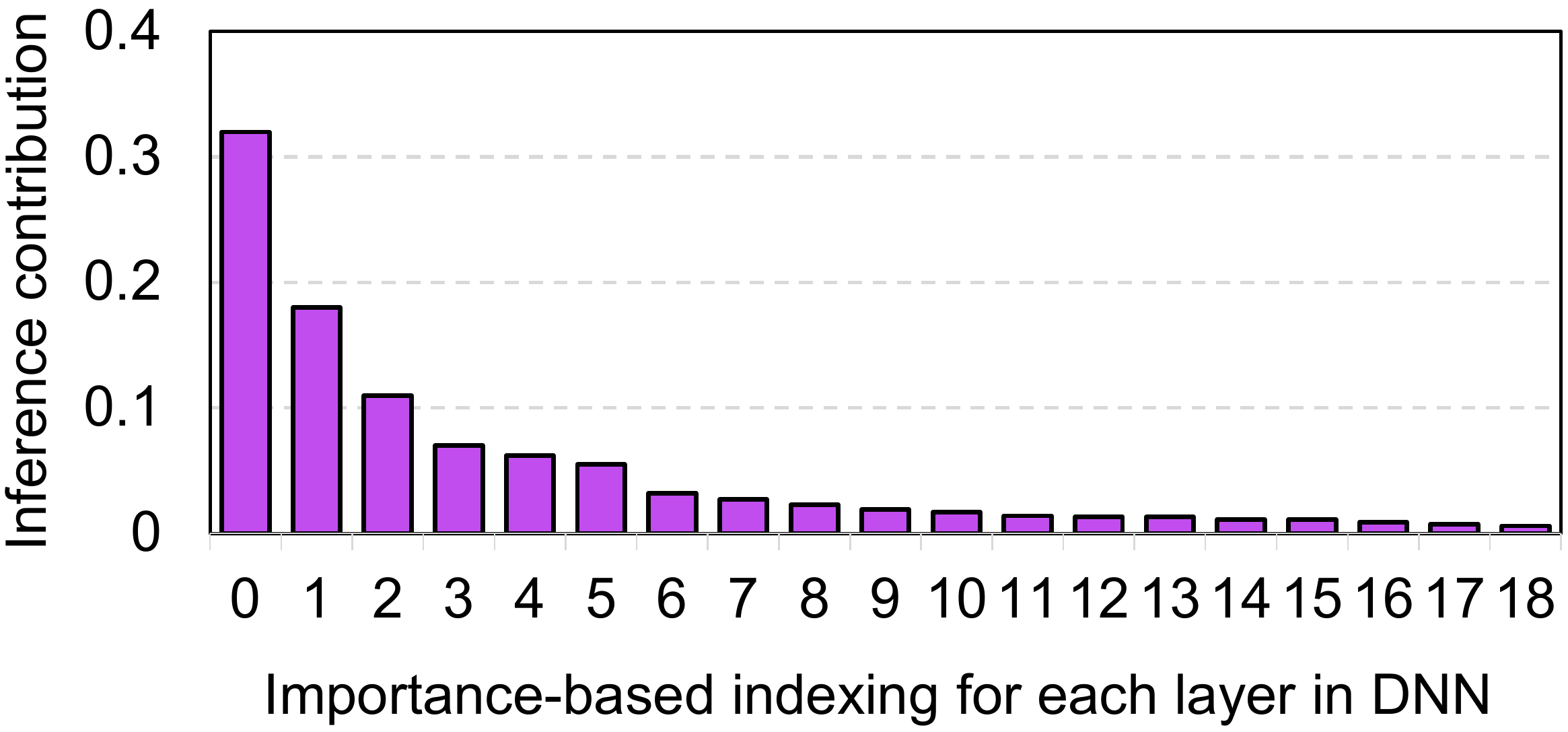}}
\caption{Descending inference contribution of each layer in ResNet-18 for CIFAR-100~\cite{krizhevsky2009learning} dataset.}
\label{fig:importance}
\end{figure}

Fig.~\ref{fig:importance} illustrates the descending inference contribution of each layer in ResNet-18 for CIFAR-100~\cite{krizhevsky2009learning} dataset, which evaluated by SCAM.
Intuitively, only a few features make major contributions to DNN inference (e.g., top-3 features of primary importance dominate 60\% of importance for the whole DNN feature maps), while a large number of secondary-importance features contributing insignificantly to DNN inference.
In this way, we can evaluate the importance of different features and keep the top-k features with primary-importance for edge execution, while the remaining secondary-importance features are compressed, and then offloaded for remote execution.
Note that compared with other explainable AI (XAI) approaches (e.g., CAM~\cite{zhou2016learning}, Grad-CAM~\cite{selvaraju2017grad}, etc.), SCAM is a lightweight and general module that can be seamlessly integrated into DNN architecture with negligible overhead and trained end-to-end together with DNN models.

In addition, offloading secondary-importance feature maps is also a challenge especially with low edge-cloud network bandwidth. 
Motivated by SPINN~\cite{laskaridis2020spinn}, we introduce precision quantization (i.e., convert the feature maps with 32-bit floating-point numbers to 8-bit fixed-length numbers) that compress secondary-importance feature maps to further reduce transmission latency. 
In this way, DVFO can effectively reduce the size of secondary-importance feature maps without significant information loss.

\subsection{Combining Local and Remote Inference Results}\label{combine}
As mentioned in Section~\ref{attention}, DVFO leverages a \emph{spatial-channel attention} mechanism to infer feature maps with primary features on edge devices, while cloud servers infer the remaining feature maps with secondary features. 
In order to efficiently and accurately fuse the inference results of both edge devices and cloud servers, DVFO applies weighted summation to fuse the inference results, and produces the final inference output at edge devices locally.

Weighted summation in DVFO we used has the following advantages, compared to neural network-based prior work such as adding an extra convolutional layer for fusion~\cite{yao2020deep}. 
First, the inference outputs of edge devices and cloud servers always maintain the same dimension. 
In contrast, using neural network (NN) layers (e.g., a fully connected or convolutional layer) to fuse these two outputs could possibly break such data alignment, hence reducing the accuracy of the final inference. 
Second, such lightweight point-to-point weighted sum has less computation than neural networks, and adds negligible overhead relative to the inference at edge devices locally.
In addition, we evaluate in detail the effect of weighted summation on accuracy and energy consumption in Section~\ref{user}.

\section{Performance Evaluation} \label{evaluation}
\subsection{DVFO Implementation} \label{implementation}
We implement offline training in DVFO with a concurrent environment~\cite{xiao2019thinking} in PyTorch, and we convert the local DNN from a float-32 model into an int-8 model using quantization aware training (QAT) supported by PyTorch. 
Different from post training dynamic quantization (PTDQ) and post training static quantization (PTSQ), QAT turns on the quantization function during the training process. 
Since quantization essentially converts the high precision of the model into low precision, which is likely to cause model performance degradation. 
In this case, QAT is better than PTDQ and PTSQ.
In addition, both the network and target network with the prioritized experience replay and $\epsilon$-greedy policy in DRL are trained using Adam optimizer. 
Each network has three hidden layers and one output layer, and each hidden layer has 128, 64, and 32 neural network units, respectively. 
We set the learning rate, buffer size and minibatch to $10^{-4}$, $10^{6}$ and 256, respectively.

Table~\ref{device} lists specific parameters of edge devices and cloud servers used in DVFO.
Since we set ten levels evenly between the maximum and the minimum CPU, GPU and memory frequencies of edge devices, there are a total of $10^{6}$ CPU-GPU-memory pairs.
We use nvpmodel, a performance and power management tool from NVIDIA, which support flexible hardware frequency scaling on-device.
\begin{table*}
\setlength{\abovecaptionskip}{0pt}
\setlength{\belowcaptionskip}{0pt}
    \caption{The specific parameters of the edge-cloud collaboration device.}
    \label{device}
    \centering
    \begin{tabular}{ccccccccc} \hline
         & Device  & CPU & GPU & Memory & CPU Freq. & GPU Freq. & Memory Freq.  & Max Power\\ \hline
    	\multirow{3}*{Edge} & NVIDIA Jetson Nano  & ARM Cortex-A57  & 128-core Maxwell & 4GB & 1479MHz  & 921.6MHz & 1600MHz & 10W \\
        & NVIDIA Jetson TX2 & ARM Cortex-A57  & 256-core Pascal & 8GB & 2000MHz  & 1300MHz & 1866MHz  & 15W\\
        & NVIDIA Xavier NX & Carmel ARMv8.2  & 384-core Volta  & 8GB & 1900MHz  & 1100MHz & 1866MHz  & 20W\\
    	Cloud & NVIDIA RTX 3080 & Intel Xeon 6226R & 8960-core Ampere & 128GB & 2900MHz & 1440MHz & 2933MHz  & 320W\\ \hline
    \end{tabular}
    \vspace{-0.1in}
\end{table*}

\subsection{Experiment Setup}
\subsubsection{Datasets and DNN models} We evaluate DVFO on CIFAR-100~\cite{krizhevsky2009learning} and ImageNet-2012~\cite{krizhevsky2017imagenet} datasets, respectively. 
The images with different sizes can comprehensively reflect the diversity of input data. 
Due to limited compute resources on edge devices, we set the batch size to be one for edge-cloud collaborative inference. 
We use EfficientNet-B0 and Vision Transformer (ViT-B16) to represent memory-intensive and compute-intensive DNN, respectively. 
Moreover, the remote DNN in DVFO is constructed by removing the first convolutional layer from the benchmark DNN~\cite{huang2022real}.

\subsubsection{Energy consumption measurement} As described in Section~\ref{Problem}, the overall energy consumption of edge devices incorporates computing and offloading energy consumption.
To be more specific, we use jetson-stats~\cite{jetson-stats}, an open source monitoring toolkit to periodically profile and record the overall energy consumption of edge devices in real time. 

\subsubsection{Baselines} We compare DVFO with the following four approaches. 
Note that all experimental results are averaged over the entire test dataset.
\begin{itemize}
    \item \textbf{AppealNet~\cite{li2021appealnet}}: An edge-cloud collaborative framework that decides whether the task uses a lightweight DNN model on edge devices or a complex DNN model on cloud servers by identifying the difficulty of the input data.
    \item \textbf{DRLDO~\cite{panda2022energy}}: A DVFS-aware offloading framework that automatically co-optimizes the CPU frequency of edge devices and the offloaded input data.
    \item \textbf{Cloud-only}: The whole feature maps are offloaded to cloud servers without edge-cloud collaboration inference.
    \item \textbf{Edge-only}: The whole model is executed on edge devices without edge-cloud collaboration inference.
\end{itemize}

Since AppealNet deploys DNN with different complexity at edge devices and cloud servers, respectively, we use the same DNN, including DVFO all the time, in order to make fair comparisons among different approaches. 
In addition, we use the same quantization (i.e., QAT) for AppealNet, DRLDO, and Cloud-only.
Note that all the experiments were performed on the devices listed in Table~\ref{device}.
NVIDIA Xavier NX as the default edge devices, unless otherwise mentioned. 
By default, we use $\eta=0.5$ to represent a balance between energy consumption and end-to-end latency. 
We also test $\eta$ from 0 to 1 in Section~\ref{user}.
The summation weight $\lambda$ is initialized to 0.5, and we also test $\lambda$ from 0 to 1 in Section~\ref{user}.

\subsection{Comparison of Inference Performance}
\begin{figure}[htbp]
\large
\centerline{\includegraphics[width=\linewidth]{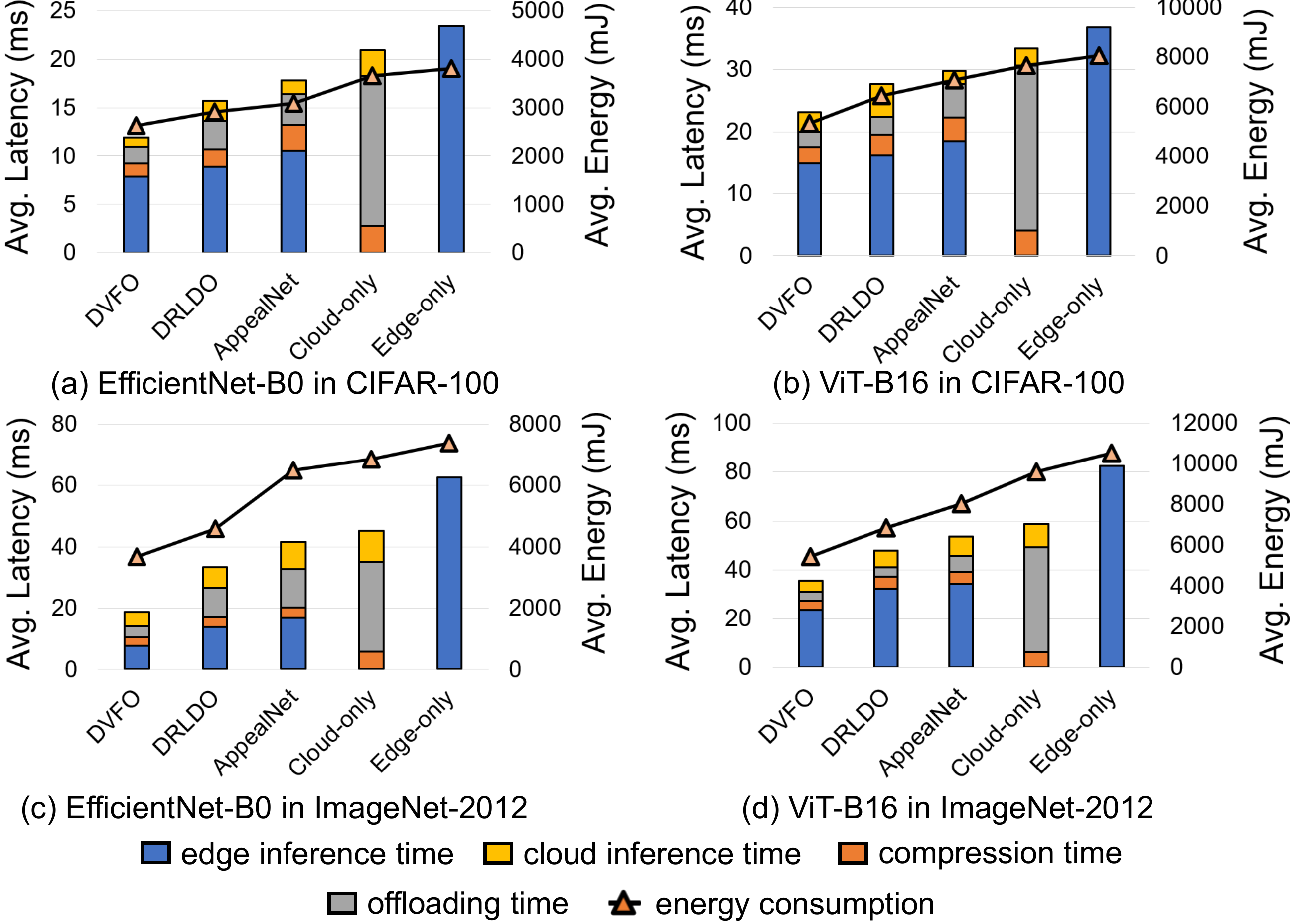}}
\caption{Comparison of end-to-end latency and energy consumption for EfficientNet-B0 and Vision Transformer using Edge-only and edge-cloud collaborative inference on CIFAR-100~\cite{krizhevsky2009learning} and ImageNet-2012~\cite{krizhevsky2017imagenet} datasets.}
\label{fig:result}
\end{figure}

We first compare the inference performance of DVFO with baselines. We use trickle, a lightweight bandwidth control suite to set the transmission rate of the network bandwidth to $5Mbps$. 
Fig.~\ref{fig:result} shows the performance comparison of EfficientNet-B0 and ViT-B16 DNN models on different datasets. 
We can see that DVFO consistently outperforms all baselines. 
To be more specific, the average energy consumption of these two DNN models using DVFO is 18.4\%, 31.2\%, 39.7\%, and 43.4\% lower than DRLDO, AppealNet, Cloud-only, and Edge-only, respectively. 
Meanwhile, DVFO significantly reduces the end-to-end latency by 28.6\%$\sim$59.1\% on average. 
Since the DNN is executed on edge devices, the end-to-end latency of Edge-only is higher than other approaches. 
Cloud-only is more sensitive to bandwidth fluctuations that leads to the highest end-to-end latency compared to other edge-cloud collaboration approaches.

Compared with DRLD and AppealNet, the reduction of energy consumption and end-to-end latency mainly have the following two aspects: 
1) \emph{Co-optimization of frequency and proportion of offloading.} DRLDO only optimizes CPU frequency and offloading proportion. 
Since DVFO also takes GPU and memory frequencies as decision variables in DRL, which can optimize the hardware frequency to further reduce energy consumption and end-to-end latency. 
In addition, AppealNet does not utilize DVFS technology to optimize frequency, therefore its energy consumption is higher than DVFO. 
More importantly, binary offloading in AppealNet is more sensitive to network bandwidth than partial offloading. 
2) \emph{Lightweight offloading mechanism.} Compared with DRLDO and AppealNet that need to offload the original feature maps to cloud servers, DVFO combines attention mechanism with quantization technology to compress the secondary-importance the feature maps to be offloaded, without losing much information.

\begin{figure}[htbp]
\large
\centerline{\includegraphics[width=\linewidth]{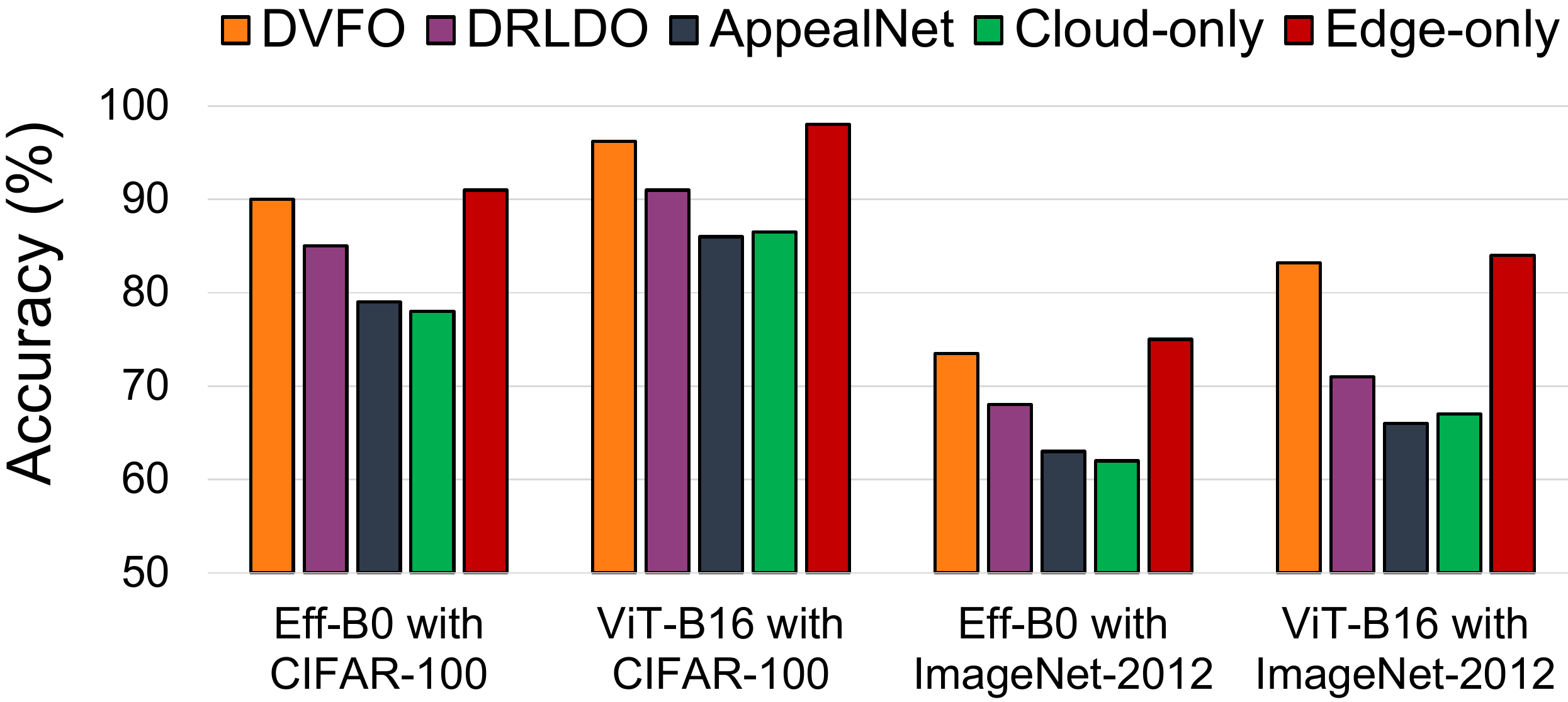}}
\caption{Comparison of benchmark DNN inference accuracy on different datasets. We set the batch size to 1.}
\label{fig:accuracy}
\end{figure}

Fig.~\ref{fig:accuracy} shows that DVFO can maintain similar inference accuracy to Edge-only (i.e., the loss of accuracy is within 2\%), compared to other baseline methods with significant drop in accuracy.
Note that Edge-only performs uncompressed original feature maps and thus achieves the highest accuracy.
Since AppealNet and Cloud-only leverage the same compression technique for binary offloading (i.e. compress the whole feature maps), they suffer from similar accuracy loss, which significantly reduces accuracy. 
DRLDO leverages the partial offloading mechanism similar to DVFO.
However, DVFO leverages a lightweight weighted summation-based fusion method, the accuracy, therefore, is higher than that of DRLDO.
Such results illustrate the effective offloading for DVFO, which utilizes an attention mechanism to identify secondary-importance features combined with high-precision quantization aware training to minimize accuracy loss.

\begin{figure}
\vspace{0pt}
\setlength{\abovecaptionskip}{0pt}
\setlength{\belowcaptionskip}{0pt}
\centering
\subfigure[EfficientNet-B0 on CIFAR-100]{
\begin{minipage}[b]{0.48\linewidth}
\includegraphics[width=1\linewidth]{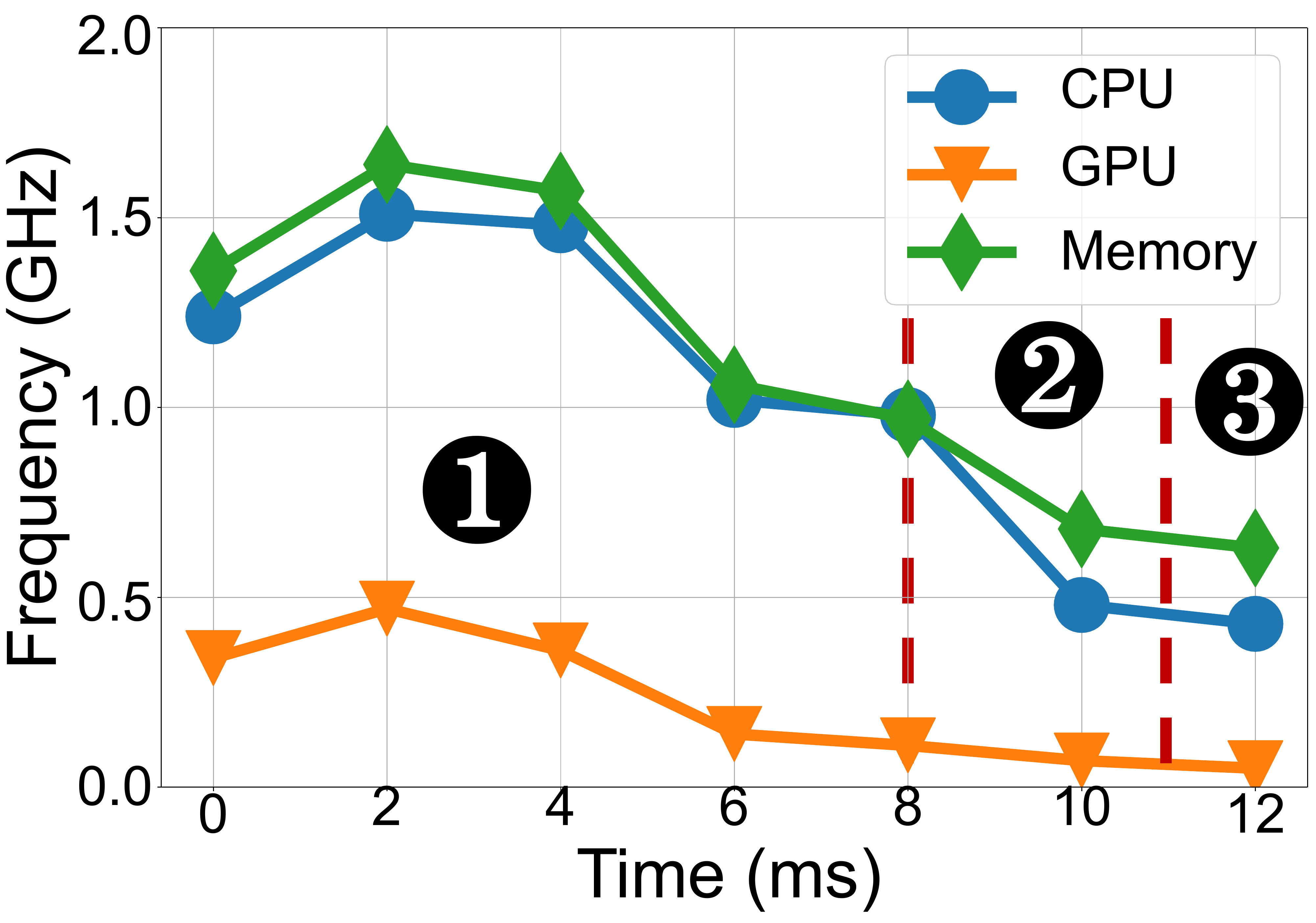}
\end{minipage}}
\subfigure[ViT-B16 on CIFAR-100]{
\begin{minipage}[b]{0.48\linewidth}
\includegraphics[width=1\linewidth]{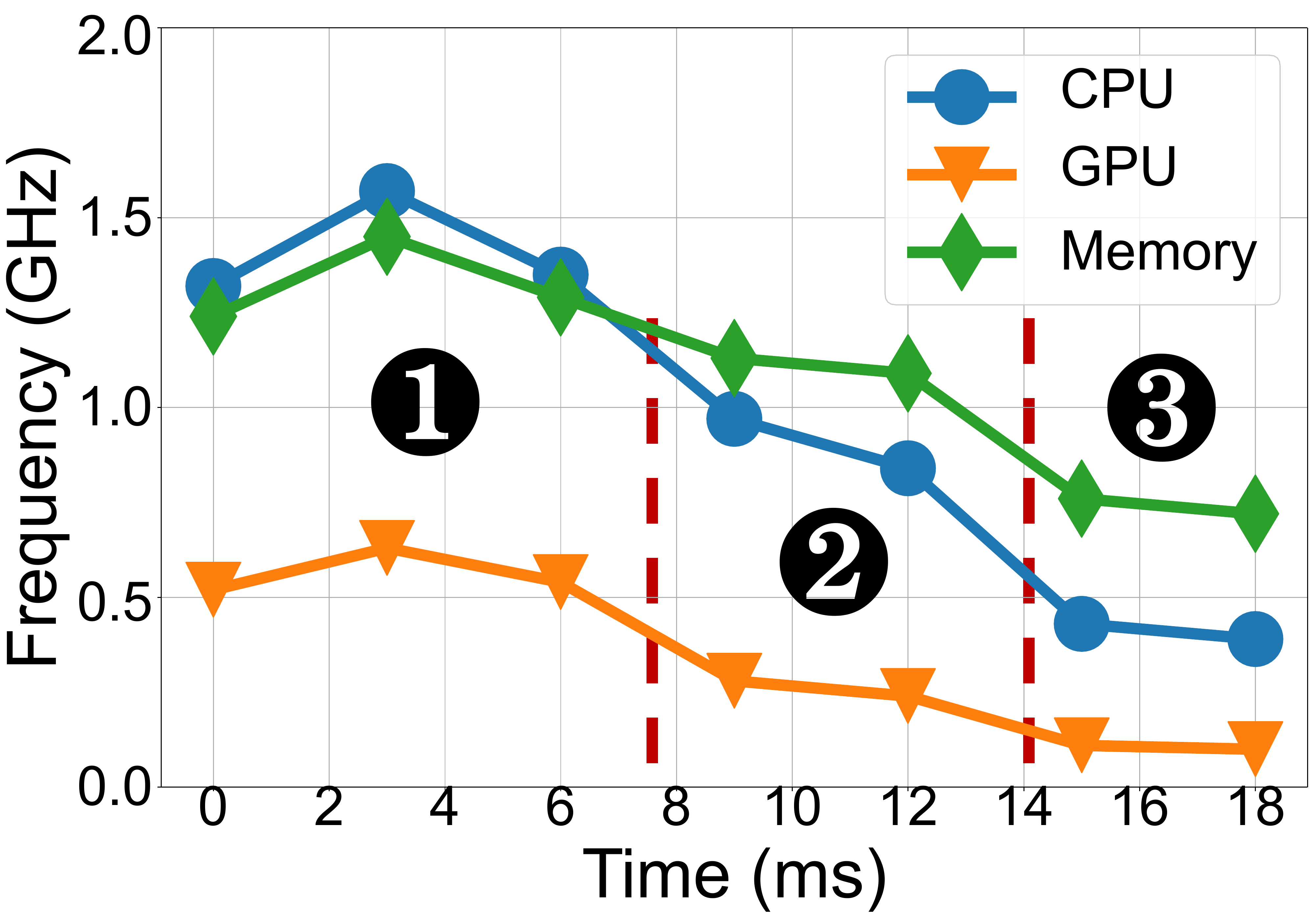}
\end{minipage}}
\subfigure[EfficientNet-B0 on ImageNet-2012]{
\begin{minipage}[b]{0.48\linewidth}
\includegraphics[width=1\linewidth]{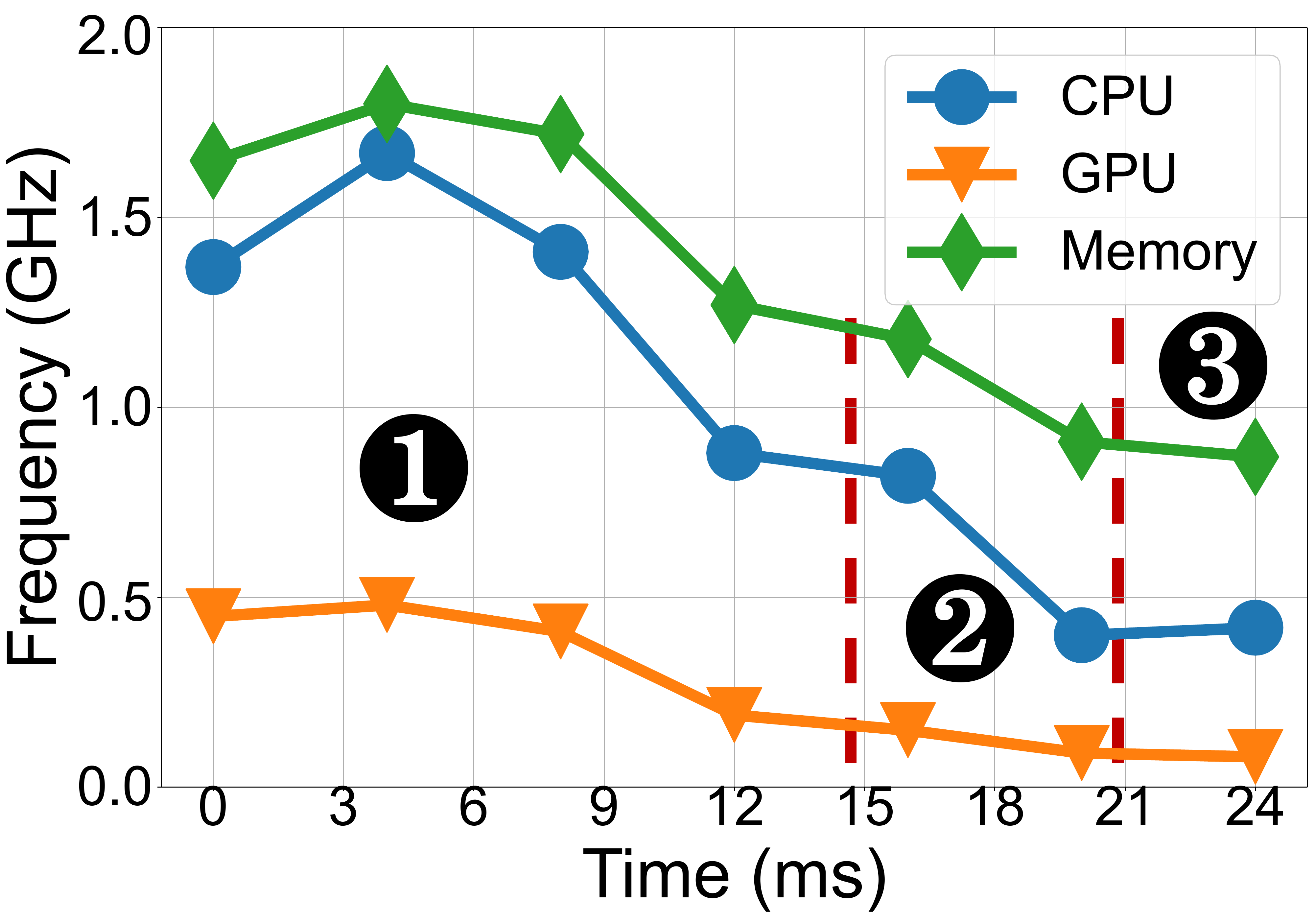}
\end{minipage}}
\subfigure[ViT-B16 on ImageNet-2012]{
\begin{minipage}[b]{0.48\linewidth}
\includegraphics[width=1\linewidth]{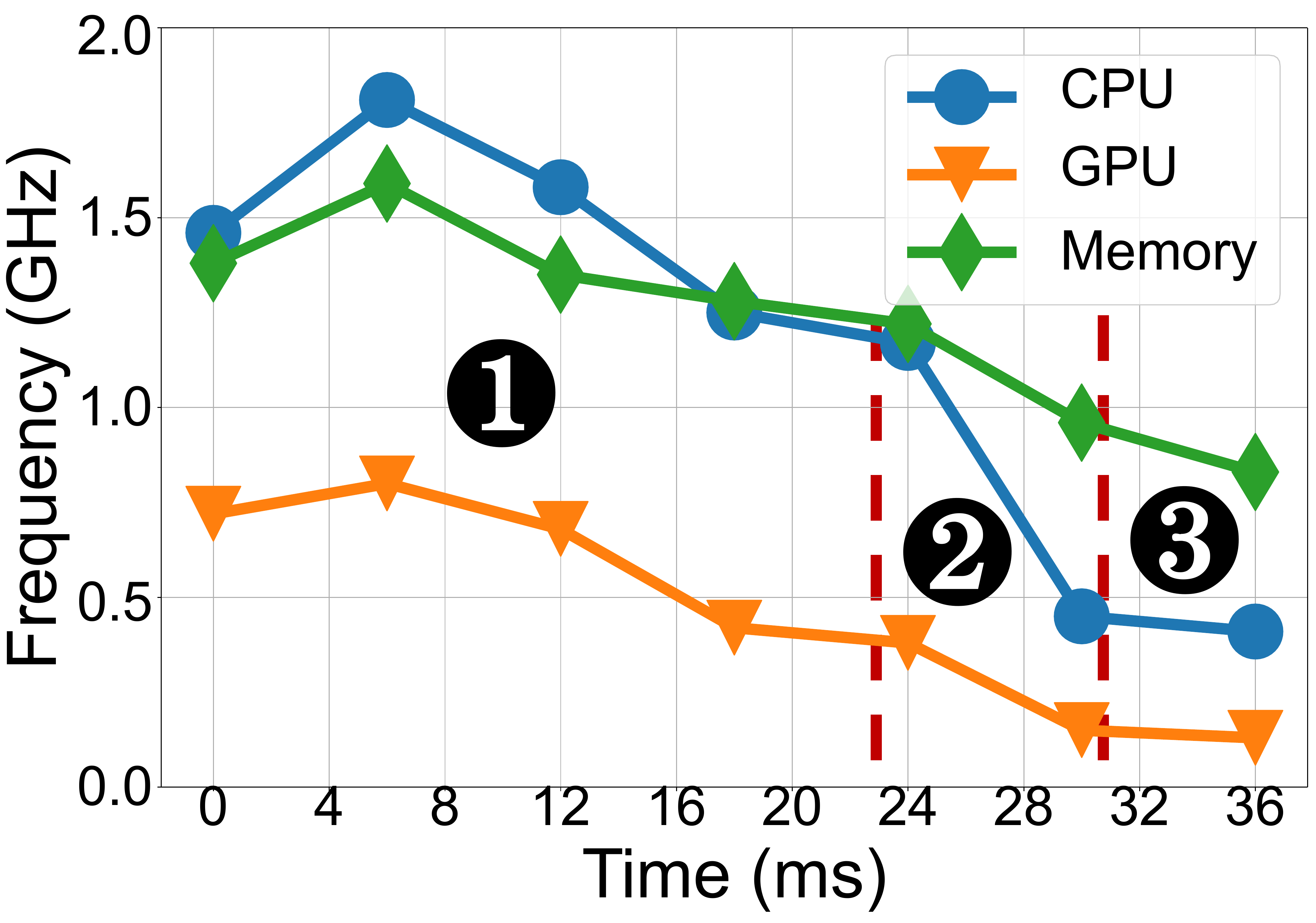}
\end{minipage}}
\caption{The trend of hardware frequencies with the execution process for EfficientNet-B0 and ViT-B16 under CIFAR-100~\cite{krizhevsky2009learning} and ImageNet-2012~\cite{krizhevsky2017imagenet} datasets, respectively. We use an NVIDIA Xavier NX edge platform as a testbad. The execution process consists of \ding{182} edge inference, \ding{183} sum of offloading and compression, and \ding{184} cloud inference (including fusion operations).}
\label{fig:frequency}
\end{figure}

In Fig.~\ref{fig:frequency}, we show the trend of the hardware frequencies with inference process on EfficientNet-B0 and ViT-B16 under different datasets, respectively. 
We take the inference of EfficientNet-B0 in Fig.~\ref{fig:frequency}(a) under CIFAR-100~\cite{krizhevsky2009learning} dataset as an example for analysis. 
The whole end-to-end latency consists of \ding{182} edge inference, \ding{183} sum of offloading and compression, and \ding{184} cloud inference (including fusion operations). 
We can conclude from Fig.~\ref{fig:observation} that EfficientNet-B0 is memory-intensive DNN model, so that the frequencies of CPU and memory dominate edge inference, while the frequency of GPU has not yet come close to the performance bottleneck. 
In contrast, the ViT-B16 DNN model with compute-intensive properties has a significant increase in the GPU frequency during edge inference (Fig.~\ref{fig:frequency}(b) and Fig.~\ref{fig:frequency}(d)), which means that ViT-B16 can effectively utilize the computing performance of the GPU.
Moreover, since DVFO adopts the attention-based lightweight compression mechanism in Section~\ref{attention} and the concurrent offloading strategy with negligible overhead (i.e., \emph{thinking-while-moving}) in Section~\ref{DRL}, the offloading and compression operations have extremely low hardware frequencies, which can save energy while reducing the offloading latency. 
For cloud inference, edge devices does not involve inference, offloading, and compression operations, so that DVFO only needs to maintain the hardware frequencies at which the system normally operates.

\subsection{Impact of Network Bandwidth}
Since network bandwidth between edge devices and cloud servers dominates the offloading latency of feature maps, it is necessary to evaluate the robustness of DVFO with different network bandwidths. 
Due to energy and cost constraints, edge devices are equipped with \emph{WiFi} modules that have lower transmission rates compared with cloud servers. 
Here we limit the network bandwidth between 2Mbps and 8Mbps to simulate different network conditions. 
\begin{figure}
\vspace{0pt}
\setlength{\abovecaptionskip}{0pt}
\setlength{\belowcaptionskip}{0pt}
\centering
\subfigure[CIFAR-100]{
\begin{minipage}[b]{0.48\linewidth}
\includegraphics[width=1\linewidth]{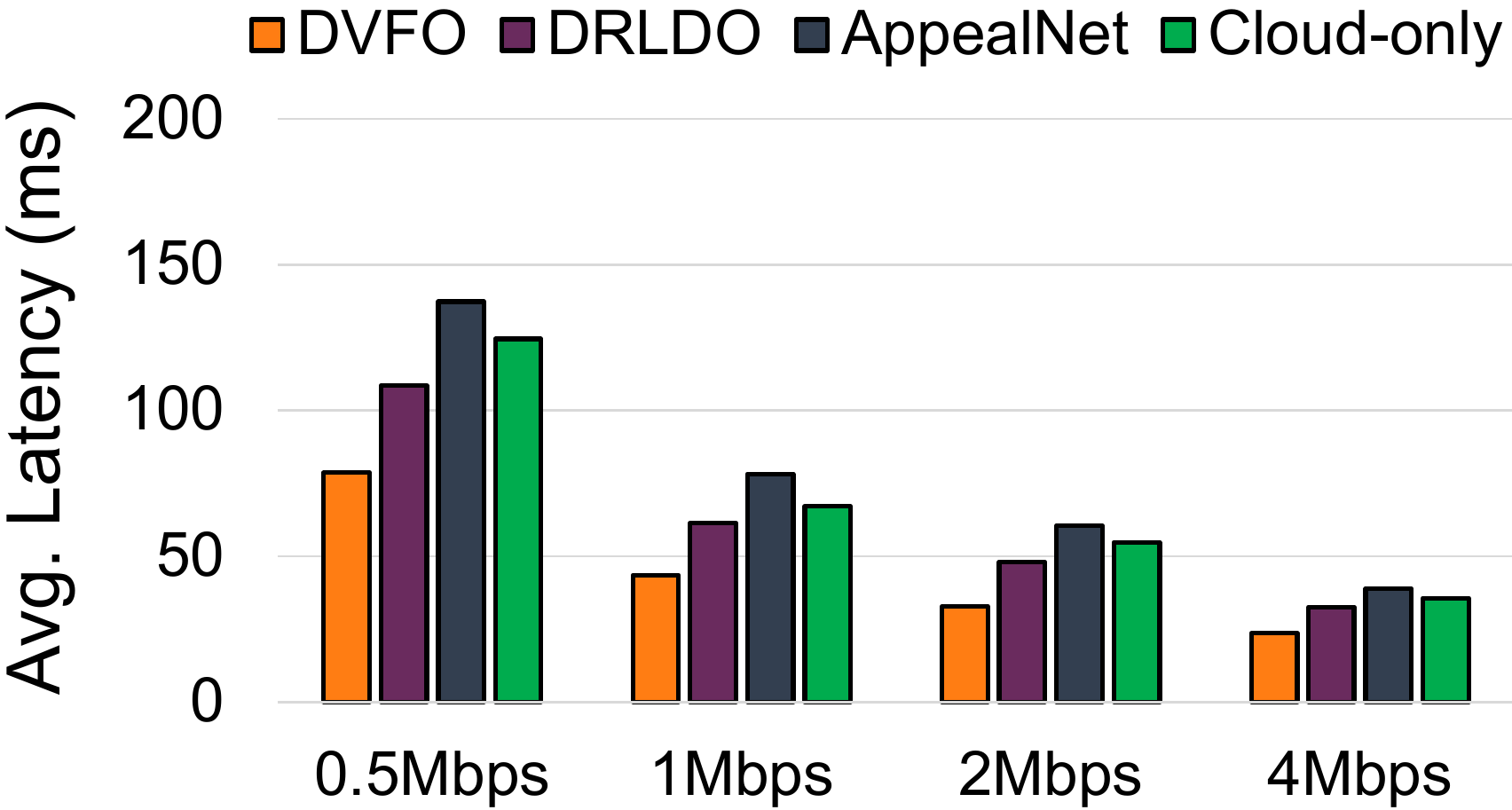}
\end{minipage}}
\subfigure[ImageNet-2012]{
\begin{minipage}[b]{0.48\linewidth}
\includegraphics[width=1\linewidth]{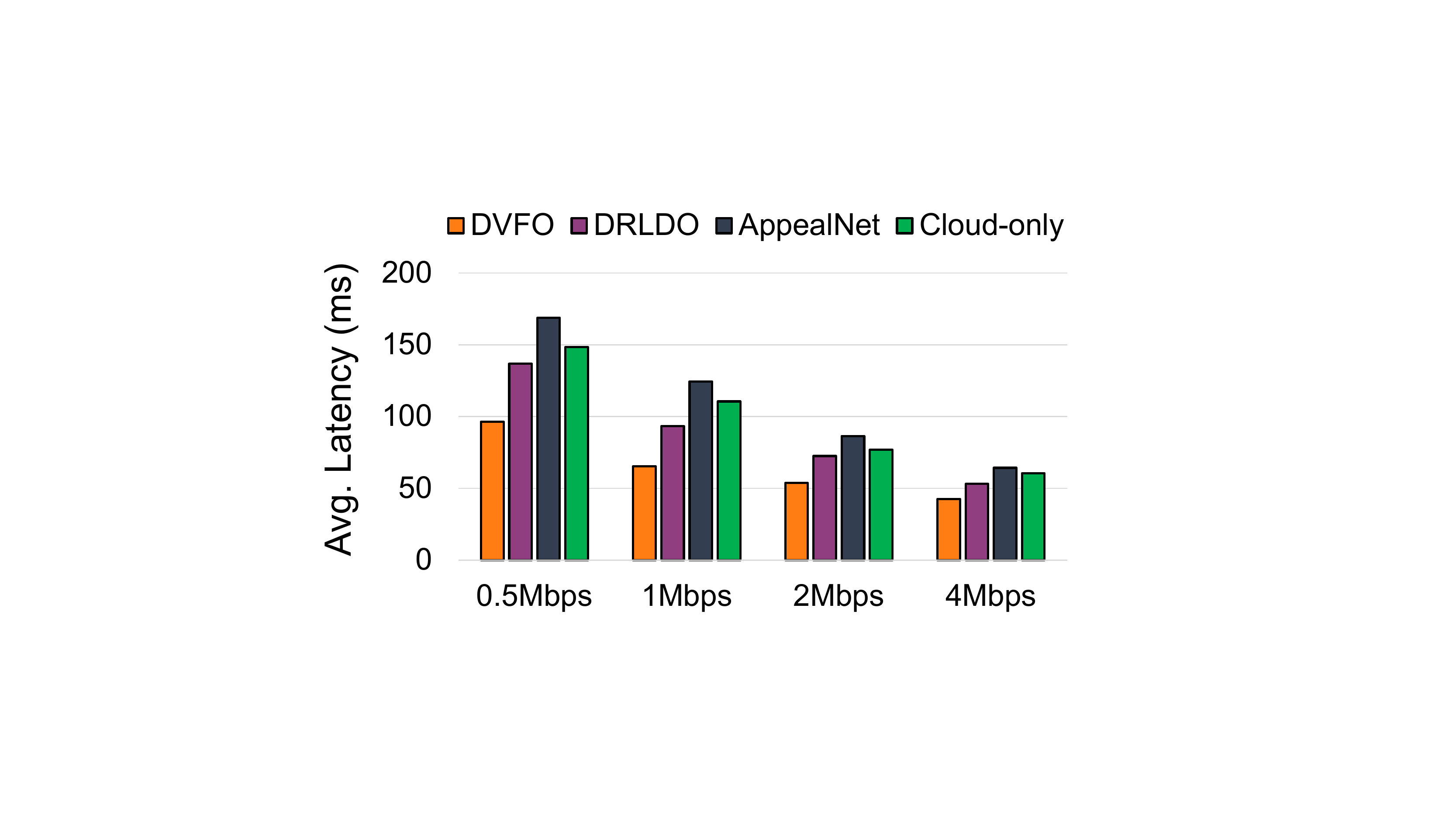}
\end{minipage}}
\caption{The end-to-end latency of EfficientNet-b0 for CIFAR-100~\cite{andrychowicz2020learning} and ImageNet-2012~\cite{krizhevsky2017imagenet} datasets under different network bandwidths.}
\label{fig:bandwidth}
\end{figure}

The results in Fig.~\ref{fig:bandwidth} illustrate that the end-to-end latency of EfficientNet-b0 with various edge-cloud collaboration approaches under CIFAR-100~\cite{andrychowicz2020learning} and ImageNet-2012~\cite{krizhevsky2017imagenet} datasets reduces with increasing network bandwidth.
It means that these approaches are limited by poor communication conditions in the case of low network bandwidth, which are more prefer to inference DNN on edge devices.
Benefit from compressing secondary-importance the partial feature maps, the end-to-end latency of DVFO is lower than other baselines, even if the available network bandwidth is only 0.5Mbps, which can effectively reduce the end-to-end latency by 28\%$\sim$43\%. 
We also observe that the performance improvement of DVFO decreases when the network bandwidth increases. 
It means that the network bandwidth is no longer a bottleneck. 
The performance improvement is mainly the appropriate adjustment of the hardware frequency for edge devices. 

In contrast, the performance of three baselines are highly dependent on network bandwidth.
On the one hand, the binary offload mechanism of ApplealNet and Cloud-only offloads all data to the cloud servers. 
In particular, the hard-case discriminator of ApplealNet adds additional overhead compared to Cloud-only, which has the highest end-to-end latency, and Cloud-only takes second place.
While DRLDO offloads part of the input data to cloud servers, the original data is not compressed. 
In addition, the think-while-moving concurrent offload mechanism in DVFO is faster than the conventional reinforcement learning-based offloading method in DRLDO, and thus has the lowest end-to-end latency.
Overall, DVFO can make better adaptive adjustments to proportion of offloading and the hardware frequency of edge devices with the fluctuation of network bandwidth. 

\subsection{Sensitivity Analysis}\label{user}
\subsubsection{Impact of the summation weights $\lambda$}
\begin{figure}
\vspace{0pt}
\setlength{\abovecaptionskip}{0pt}
\setlength{\belowcaptionskip}{0pt}
\centering
\subfigure[CIFAR-100]{
\begin{minipage}[b]{0.48\linewidth}
\includegraphics[width=1\linewidth]{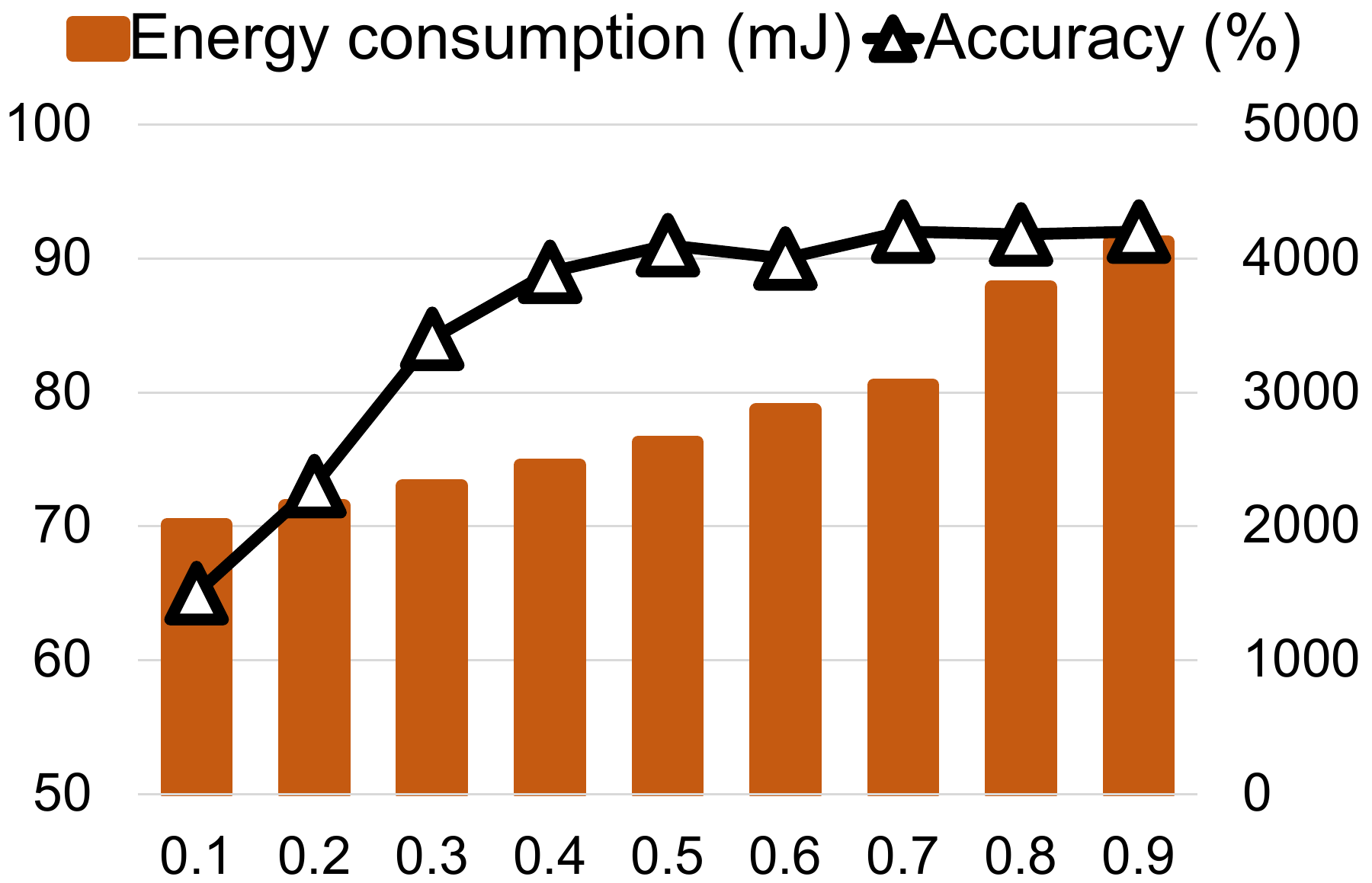}
\end{minipage}}
\subfigure[ImageNet-2012]{
\begin{minipage}[b]{0.48\linewidth}
\includegraphics[width=1\linewidth]{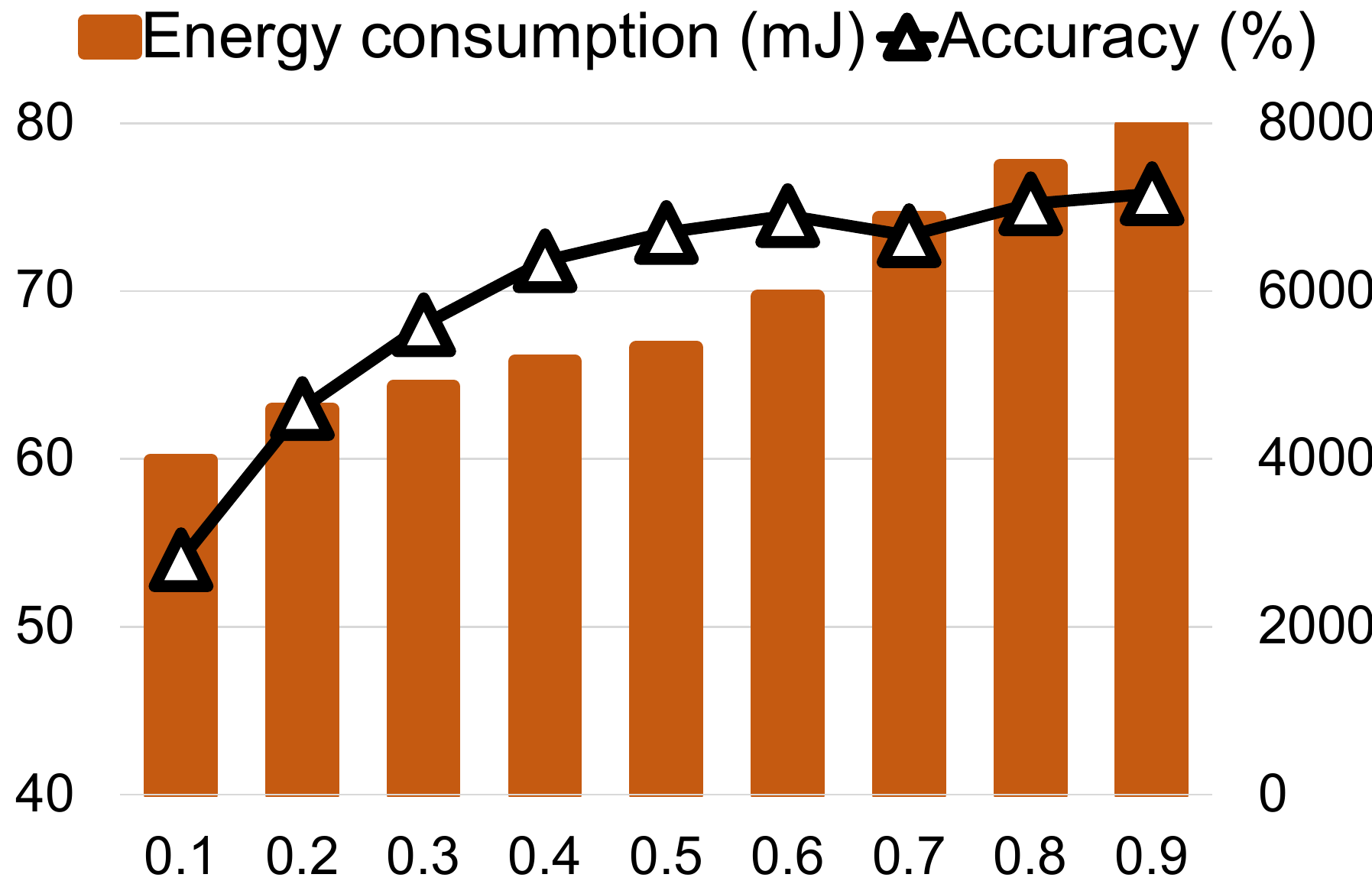}
\end{minipage}}
\caption{Sensitivity analysis of summation weights $\lambda$ on different datasets. We use EfficientNet-b0~\cite{tan2019efficientnet} as a test case.}
\label{fig:lambda}
\end{figure}

Taking EfficientNet-b0~\cite{tan2019efficientnet} as an example, Fig.\ref{fig:lambda}(a) and Fig.\ref{fig:lambda}(b) shows the impact of the summation weight $\lambda$ on performance for CIFAR-100~\cite{andrychowicz2020learning} and ImageNet-2012~\cite{krizhevsky2017imagenet} datasets, respectively. 
It can be seen that energy consumption and inference accuracy improve with the increase of $\lambda$. In particular, a smaller $\lambda$ ($\leq0.2$) significantly decreases accuracy, while a higher $\lambda$ ($\geq0.8$) sharply increases inference energy consumption. 
The intuition behind this is that a smaller $\lambda$ reduces the contribution of important features locally, which misses some important information in inference and degrades accuracy.
In contrast, increasing the value of $\lambda$, forces the majority of the inference tasks to the local DNN, which leads to higher energy consumption. 
Note that the optimal value of $\lambda$ depends on the characteristics of the data in the training dataset. 
In practice, setting $\lambda$ to an appropriate value between 0.4 and 0.6 can effectively reduce energy consumption while maintaining high-accuracy.

\subsubsection{Impact of the relative importance coefficient $\eta$}
\begin{figure}
\vspace{0pt}
\setlength{\abovecaptionskip}{0pt}
\setlength{\belowcaptionskip}{0pt}
\centering
\subfigure[CIFAR-100]{
\begin{minipage}[b]{0.48\linewidth}
\includegraphics[width=1\linewidth]{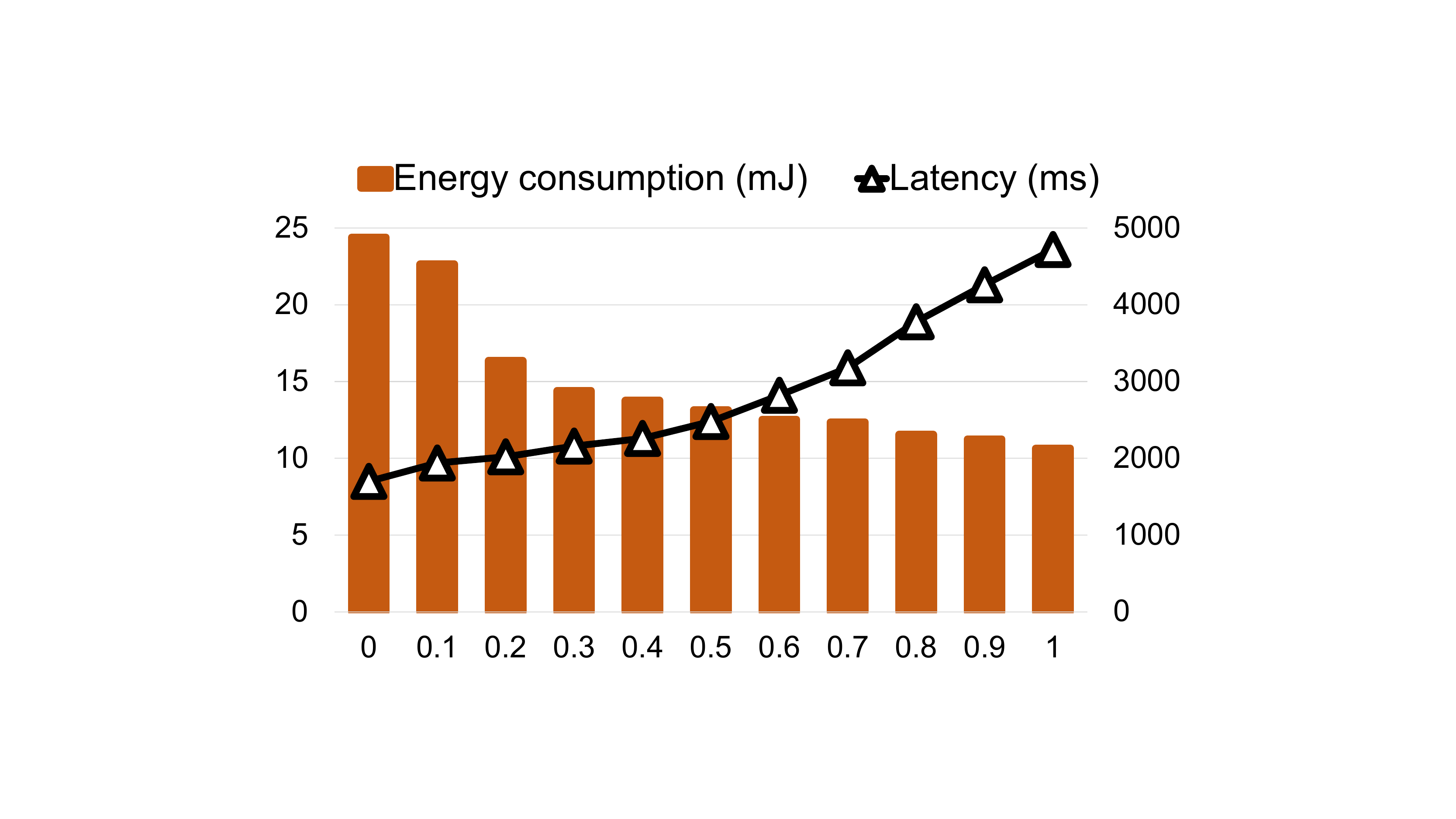}
\end{minipage}}
\subfigure[ImageNet-2012]{
\begin{minipage}[b]{0.48\linewidth}
\includegraphics[width=1\linewidth]{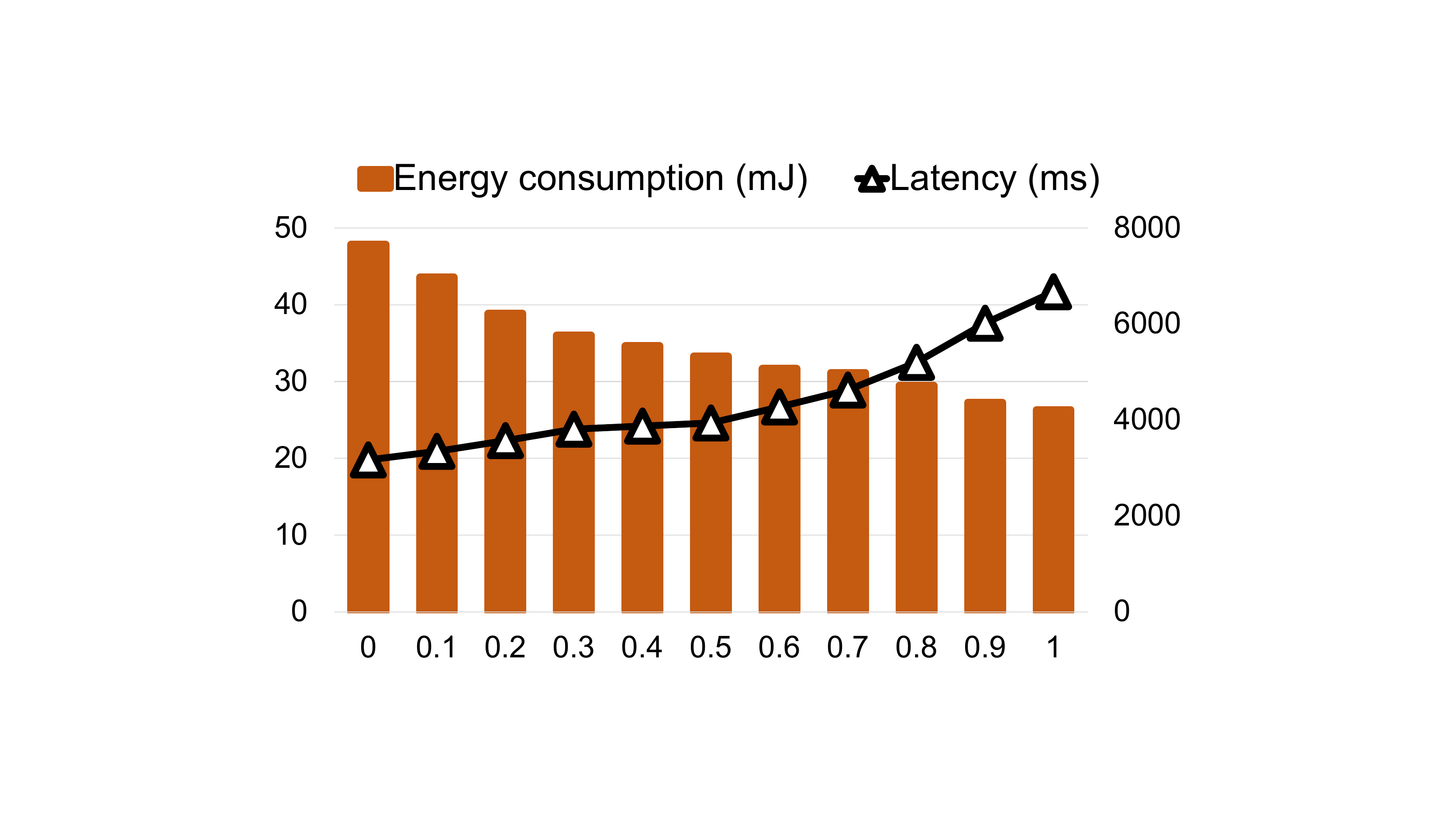}
\end{minipage}}
\caption{Sensitivity analysis of weight parameter $\eta$ on different datasets. 
We use EfficientNet-b0~\cite{tan2019efficientnet} as a test case.}
\label{fig:eta}
\end{figure}

In Fig.\ref{fig:eta}, we also take EfficientNet-b0~\cite{tan2019efficientnet} as an example to show the impact of weight parameter $\eta$ that trade-off between energy consumption and end-to-end latency, given different datasets.
We observe that DVFO significantly reduces energy consumption with increasing values of $\eta$ ($\geq0.3$), improving energy efficiency while maintaining low end-to-end latency $\eta$ ($\leq0.6$). 
Specifically, compared to $\eta=0.1$, even though DVFO reduced end-to-end latency by up to 39.2\% at $\eta=0.4$ on CIFAR-100~\cite{krizhevsky2009learning} dataset, the end-to-end latency is only increased by 16.5\%. 
We also observe a similar phenomenon on the ImageNet-2012~\cite{krizhevsky2017imagenet} dataset.
In summary, DVFO allows users to adjust the trade-off between energy consumption and end-to-end latency by selecting an appropriate weight parameter $\eta$.

\subsection{Comparison of various fusion methods}\label{fuse}
In this section, we compare the accuracy loss and runtime overhead (i.e.,  energy consumption and end-to-end latency) induced by weighted summation in DVFO and NN-based fusion methods, compared to single-device inference (without fusion). 
In particular, it can be seen from Fig.~\ref{fig:lambda} that the appropriate value of $\lambda$ has different preferences under different datasets. 
To achieve high accuracy while maintaining low energy consumption, we set $\lambda$ to 0.5 for CIFAR-100~\cite{andrychowicz2020learning} and 0.6 for ImageNet-2012~\cite{krizhevsky2017imagenet}, respectively.
We use a filter size of 3$\times$3 and a softmax function to implement a convolutional layer and a fully connected layer for fusing inference results, respectively.

As shown in Table~\ref{fusion}, the weighted summation we used in DVFO can achieve the lowest accuracy loss (within 1\%), compared to single-device inference. 
This is because weighted summation enables both the inference results of edge devices and cloud servers to maintain a highly consistent data alignment, and such lightweight point-to-point weighting has low-complexity with negligible overhead. 
The main challenge of weighted summation is that the output of Local DNN at edge devices and Remote DNN at cloud servers may potentially have a big difference. 
For instance, a few but the output values with primary importance in Local DNN could be overlapped by the the output values with secondary importance in Remote DNN. 
We can maintain $\lambda$ in an appropriate range by manual fine-tuning or utilizing a learning-based adaptive strategy in DVFO, thereby minimizing the additional inference accuracy loss that may result.
In contrast, neural network-based fusion approaches (i.e., fully connected layers and convolutional layers) have significant accuracy loss, which is 6.7$\times$ and 12.3$\times$ that of the weighted summation in DVFO, respectively. 
As pointed out in Section~\ref{combine}, neural network-based fusion approaches break the alignment of weighted values and thus significantly reduce inference accuracy.
\begin{table}
\setlength{\abovecaptionskip}{0pt}
\setlength{\belowcaptionskip}{0pt}
    \caption{Comparison of fuse methods.}
    \label{fusion}
    \centering
    \begin{tabular}{l|cc} \hline 
    	\multirow{2}{*}{Fusion methods} & \multicolumn{2}{c}{Accuracy (\%)} \\ 
                              & CIFAR-100               & ImageNet-2012 \\ \hline
    	Single-device (without fusion) & 91.84                   & 74.52   \\ 
    	Fully connected-based NN layer & 87.39 (\textbf{4.45} $\downarrow$)  & 70.63 (\textbf{3.89} $\downarrow$)  \\
    	Convolutional-based NN layer   & 82.93 (\textbf{8.91} $\downarrow$)  & 68.24 (\textbf{6.28} $\downarrow$)  \\ 
    	\textbf{DVFO (Ours)}  & \textbf{91.16} (\textbf{0.68} $\downarrow$) & \textbf{73.96} (\textbf{0.56} $\downarrow$)      \\ \hline
    \end{tabular}
\end{table}

\begin{figure}
\vspace{0pt}
\setlength{\abovecaptionskip}{0pt}
\setlength{\belowcaptionskip}{0pt}
\centering
\subfigure[Energy consumption]{
\begin{minipage}[b]{0.48\linewidth}
\includegraphics[width=\linewidth]{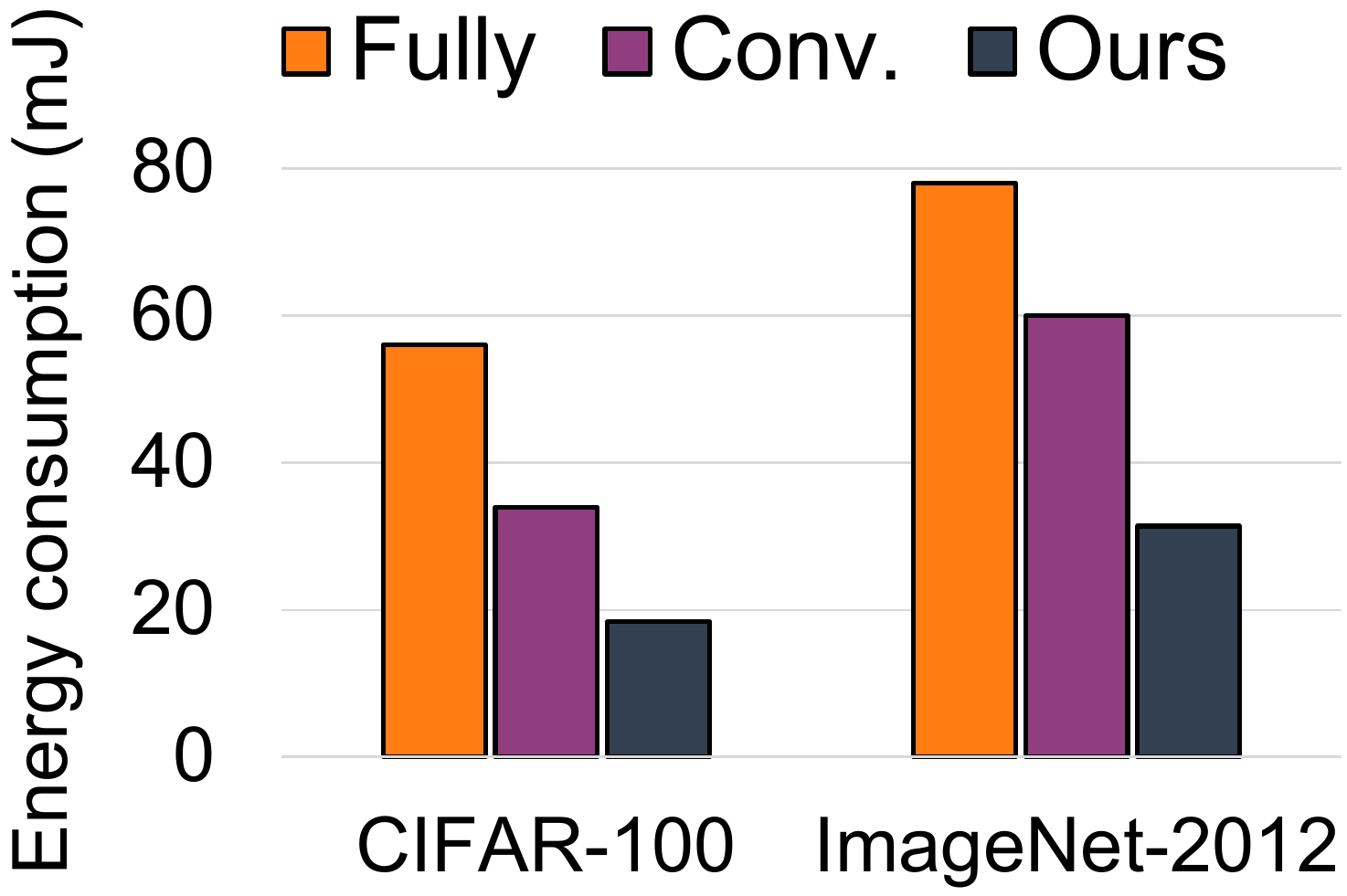}
\end{minipage}}
\subfigure[End-to-end latency]{
\begin{minipage}[b]{0.47\linewidth}
\includegraphics[width=\linewidth]{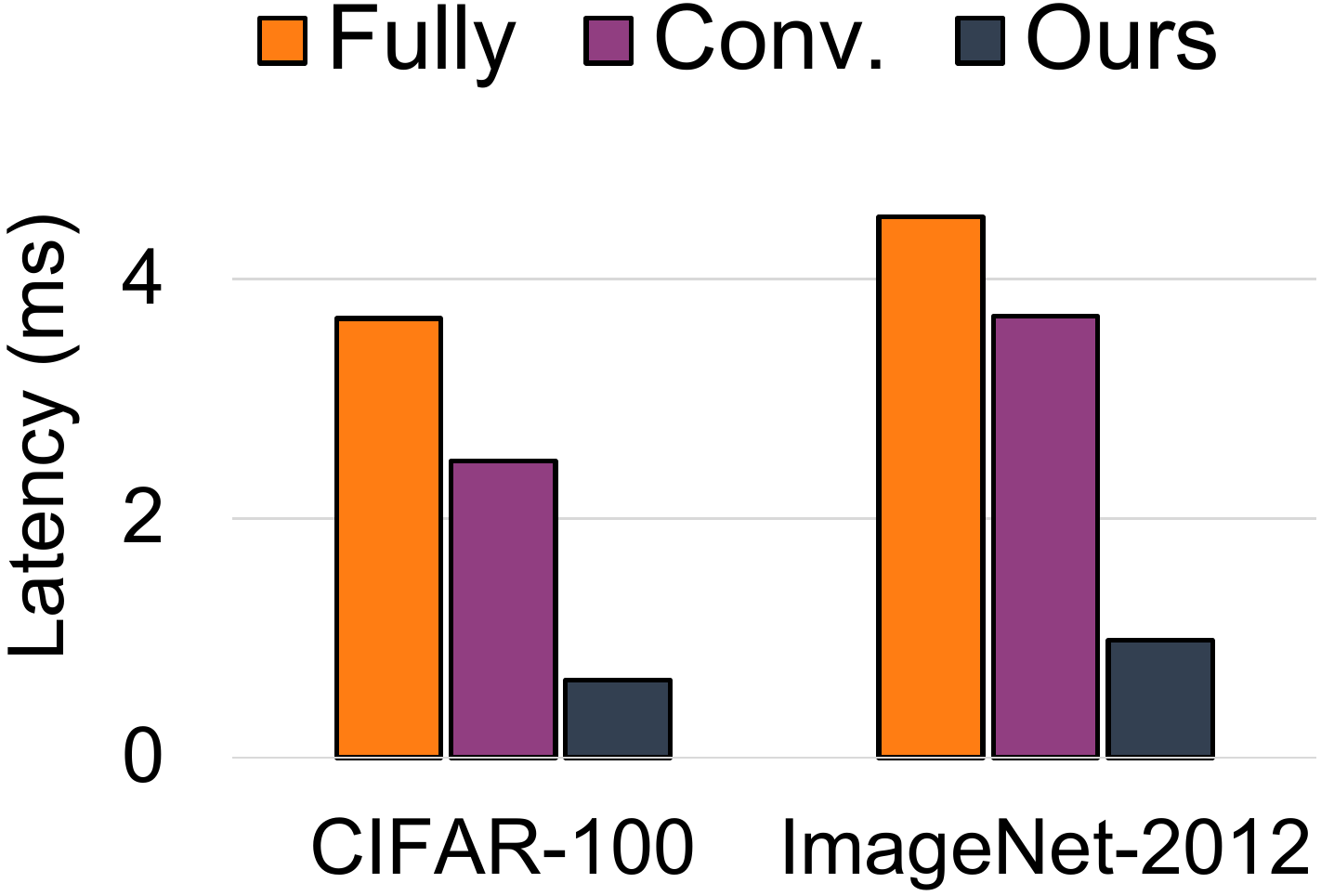}
\end{minipage}}
\caption{Comparison of runtime overhead for different fuse methods.}
\label{fig:overhead_fuse}
\end{figure}

As shown the result in Fig.~\ref{fig:overhead_fuse}, we compare the runtime overhead of weighted summation and NN-based fusion methods (i.e. convolutional and fully connected layers). 
First, in terms of energy consumption in Fig.~\ref{fig:overhead_fuse}(a), compared to the NN-based fusion method, the energy consumption of weighted summation is reduced by 56.8\% on average. 
Second, weighted summation reduces the average end-to-end latency by up to 77.5\%, as shown in Fig.~\ref{fig:overhead_fuse}(b).
It also illustrates the high energy efficiency of such lightweight point-to-point fusion methods.
In contrast, the NN-based fusion methods significantly reduce energy efficiency due to inherently expensive computation.

\subsection{Overhead Analysis}
\subsubsection{Training Overhead}
\begin{figure}[htbp]
\large
\centerline{\includegraphics[width=0.75\linewidth]{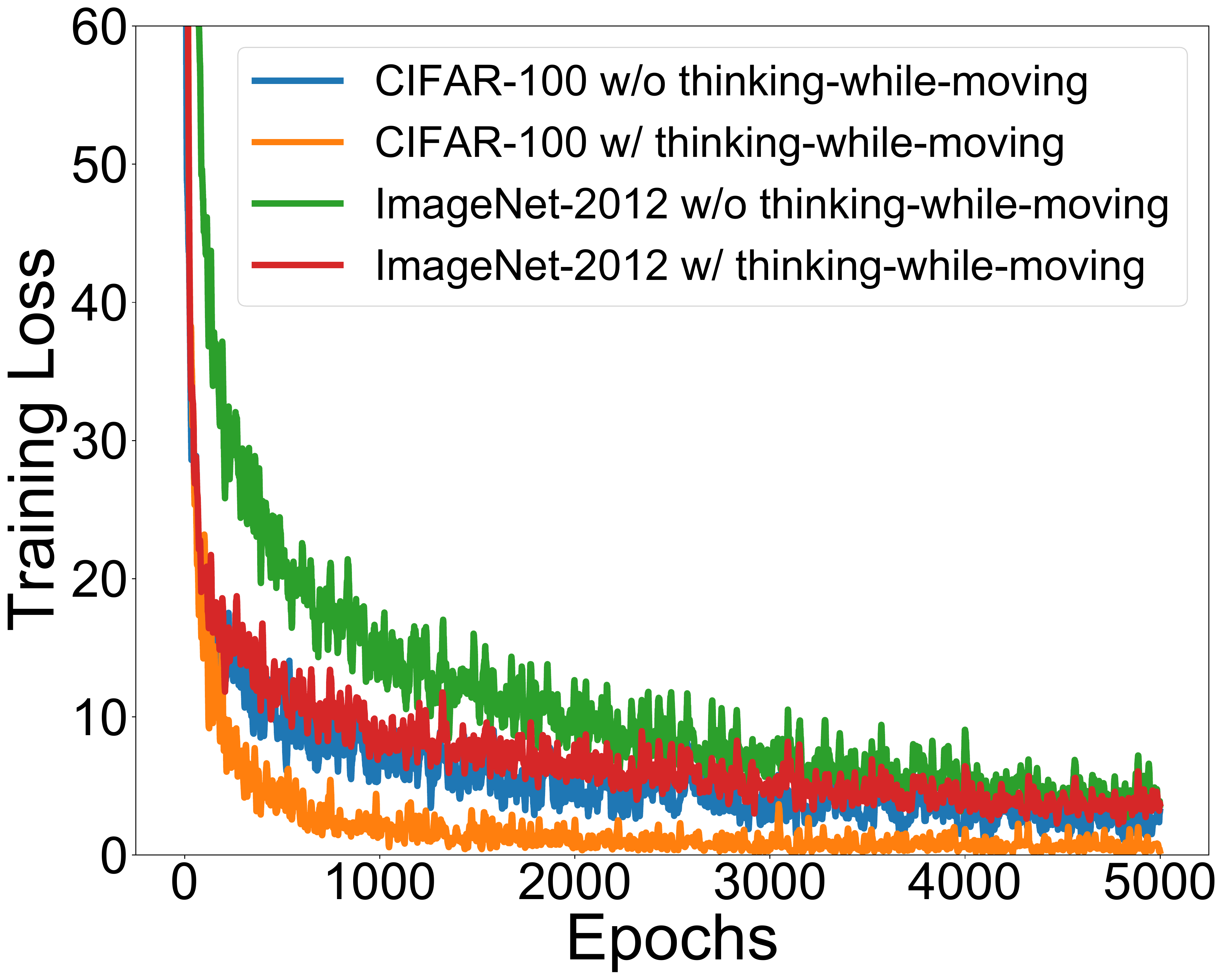}}
\caption{The training performance of DVFO with/without a \emph{thinking-while-moving} mechanism on CIFAR-100~\cite{andrychowicz2020learning} and ImageNet-2012~\cite{krizhevsky2017imagenet} datasets. We use EfficientNet-b0~\cite{tan2019efficientnet} as a test case.}
\label{fig:cost}
\end{figure}

We first evaluate the training overhead comparison of DVFO with/without a \emph{thinking-while-moving} training strategies, here we use EfficientNet-b0~\cite{tan2019efficientnet} on CIFAR-100~\cite{andrychowicz2020learning} and ImageNet-2012~\cite{krizhevsky2017imagenet} datasets as a test case.
As shown in Fig.~\ref{fig:cost}, DVFO with \emph{thinking-while-moving} shows faster convergence during the training procedure, indicating that although the attention module increases the complexity of learning, DVFO can still guarantee fast convergence of training by designing appropriate cost metrics and parallel strategies.

\subsubsection{Runtime Overhead}
\begin{figure}
\vspace{0pt}
\setlength{\abovecaptionskip}{0pt}
\setlength{\belowcaptionskip}{0pt}
\centering
\subfigure[CIFAR-100]{
\begin{minipage}[b]{0.48\linewidth}
\includegraphics[width=\linewidth]{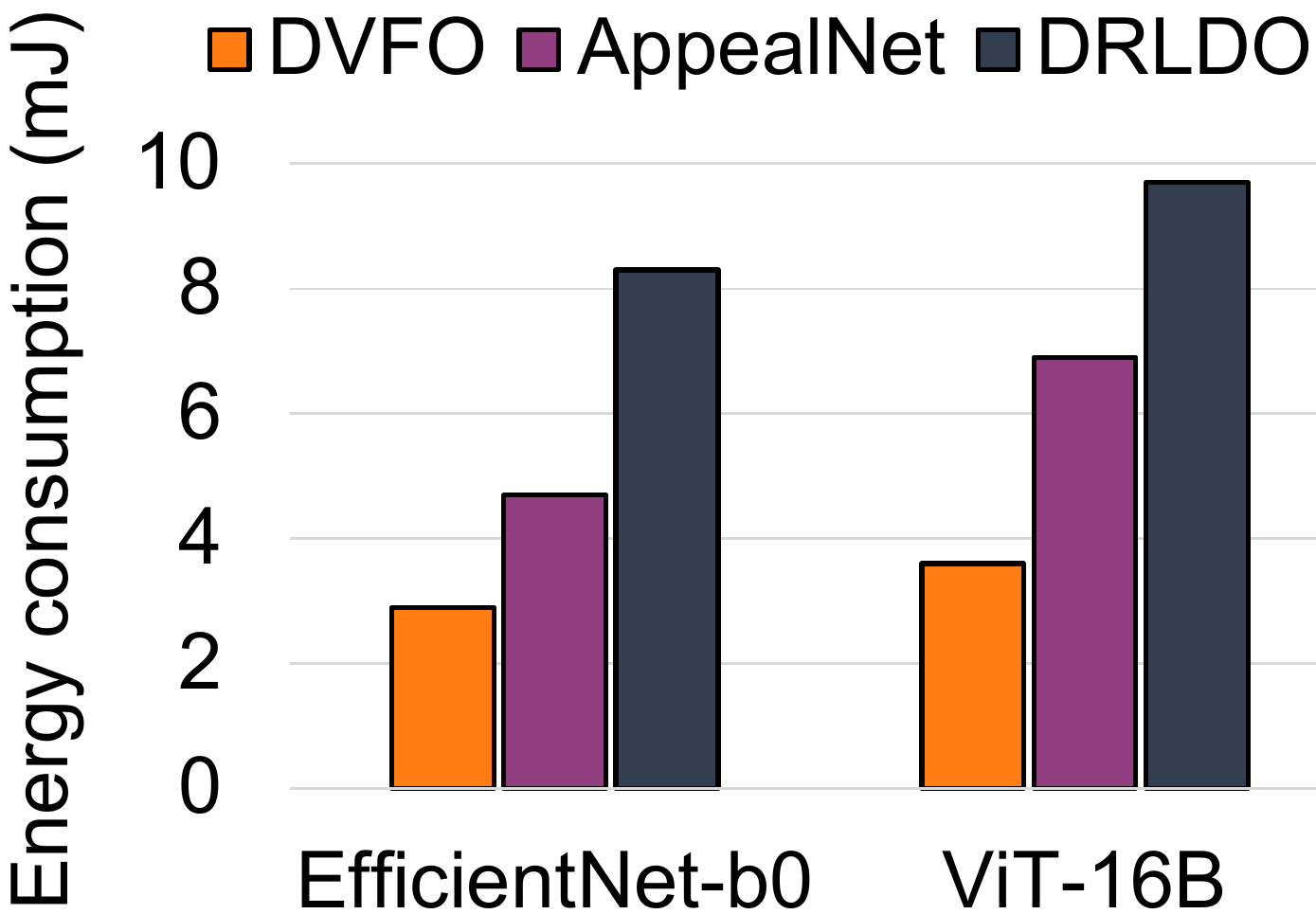}
\end{minipage}}
\subfigure[ImageNet-2012]{
\begin{minipage}[b]{0.48\linewidth}
\includegraphics[width=\linewidth]{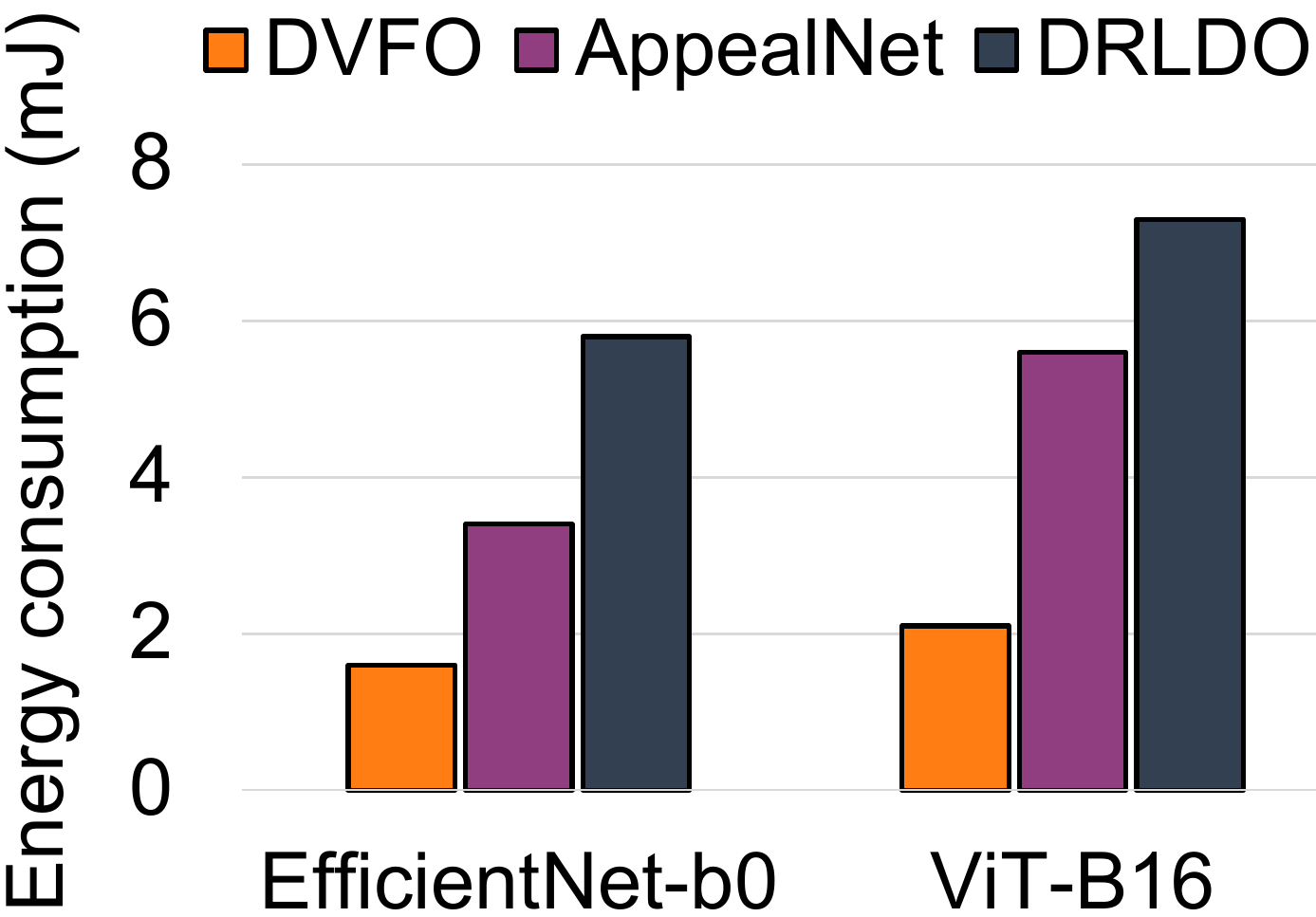}
\end{minipage}}
\caption{Comparison of runtime overhead for different datasets.}
\label{fig:overhead}
\end{figure}

The attention module in DVFO introduces additional runtime overhead.
We evaluate the energy consumption of the attention module (i.e., SCAM) averaged over 10 inference. 
As shown in Fig.~\ref{fig:overhead}, DVFO consumes less energy due to uses an extremely lightweight attention module. 
The energy consumption of DVFO is 38\%$\sim$62\% lower than AppealNet and 63\%$\sim$71\% lower than DRLDO.

\subsection{Evaluation of Scalability}
In this section, we evaluate the scalability of DVFO on other heterogeneous edge devices (i.e., NVIDIA Jetaon Nano and TX2 in Table~\ref{device}), as well as a wide range of DNN models. 
We take the widely-deployed deep learning models, ResNet-18~\cite{he2016deep}, Inception-v3~\cite{szegedy2016rethinking}, and MobileNet-v2~\cite{sandler2018mobilenetv2}, as the image classification services, the widely-deployed deep learning models, YOLOv3-Tiny~\cite{redmon2016you} and RetinaNet~\cite{lin2017focal}, as the object detection services, and the widely-deployed deep learning model, DeepSpeech~\cite{hannun2014deep}, as the speech recognition service.
The evaluation results are illustrated in Table~\ref{scale_CIFAR} and Table~\ref{scale_Image}.
It can be seen that DVFO consistently outperforms AppealNet and DRLDO in terms of end-to-end latency, energy consumption, and accuracy loss, respectively. 
Specifically, DVFO reduces the average end-to-end latency by 49.7\% and 36.2\% on Jetson Nano edgde device, compared with AppealNet and DRLDO, respectively, and has a significant performance improvement (45.4\%$\sim$63.8\%) on Jetson TX2 with abundant computing resources. 
Moreover, DVFO achieves 42.6\%$\sim$53\% energy saving on average for Jetson Nano. 
Since the basic power consumption of Jetson TX2 is higher than that of Jetson Nano, the performance improvement of energy saving is limited, but it is also outperforms the baseline (15.7\%$\sim$25.6\%). 
As mentioned in Section~\ref{fuse}, benefit from the efficient fusion method based on weighted summation, the average accuracy loss of DVFO on different datasets and heterogeneous edge devices remains within 1\%, which is much lower than that of AppealNet and DRLDO ($2.07\times$$\sim$$5.4\times$). 
Overall, DVFO can seamlessly adapt to heterogeneous edge devices and various widely-deployed DNN models, and thus it has flexible scalability.
\begin{table*}
\setlength{\abovecaptionskip}{0pt}
\setlength{\belowcaptionskip}{0pt}
    \caption{Evaluation of scalability on CIFAR-100 dataset.}
    \label{scale_CIFAR}
    \centering
    \begin{tabular}{c|c|ccc|ccc|ccc} \hline 
\multirow{2}{*}{Edge Device} & \multirow{2}{*}{Model} & \multicolumn{3}{c|}{End-to-end latency (ms)} & \multicolumn{3}{c|}{Energy consumption (mJ)} & \multicolumn{3}{c}{Accuracy loss (\%)}  \\  
            &           & \multicolumn{1}{c}{AppealNet} & \multicolumn{1}{c}{DRLDO} & \multicolumn{1}{c|}{\textbf{DVFO}} & \multicolumn{1}{c}{AppealNet} & \multicolumn{1}{c}{DRLDO} & \multicolumn{1}{c|}{\textbf{DVFO}} & \multicolumn{1}{c}{AppealNet} & \multicolumn{1}{c}{DRLDO} & \multicolumn{1}{c}{\textbf{DVFO}} \\ \cline{1-11}
\multirow{6}{*}{\makecell[c]{NVIDIA \\ Jetson Nano}} & ResNet-18   & \multicolumn{1}{c}{24.1} & \multicolumn{1}{c}{20.6} & \textbf{14.8} & \multicolumn{1}{c|}{6217} & \multicolumn{1}{c|}{5723} & \textbf{4867} & \multicolumn{1}{c|}{4.62} & \multicolumn{1}{c|}{2.13} & \textbf{0.54} \\
& Inception-v4            & \multicolumn{1}{c}{25.6} & \multicolumn{1}{c}{24.9} & \textbf{16.7} & \multicolumn{1}{c|}{6894} & \multicolumn{1}{c|}{6421} & \textbf{5346} & \multicolumn{1}{c|}{5.14} & \multicolumn{1}{c|}{3.47} & \textbf{0.98} \\
& MobileNet-v2            & \multicolumn{1}{c}{30.7} & \multicolumn{1}{c}{28.4} & \textbf{17.6} & \multicolumn{1}{c|}{7082} & \multicolumn{1}{c|}{6804} & \textbf{5623} & \multicolumn{1}{c|}{2.84} & \multicolumn{1}{c|}{1.69} & \textbf{0.63} \\
& YOLOv3-Tiny             & \multicolumn{1}{c}{26.8} & \multicolumn{1}{c}{24.7} & \textbf{18.4} & \multicolumn{1}{c|}{6489} & \multicolumn{1}{c|}{6174} & \textbf{5391} & \multicolumn{1}{c|}{3.87} & \multicolumn{1}{c|}{2.26} & \textbf{0.68} \\
& RetinaNet               & \multicolumn{1}{c}{35.4} & \multicolumn{1}{c}{31.6} & \textbf{25.9} & \multicolumn{1}{c|}{8256} & \multicolumn{1}{c|}{7498} & \textbf{6157} & \multicolumn{1}{c|}{2.65} & \multicolumn{1}{c|}{1.54} & \textbf{0.78} \\
& DeepSpeech              & \multicolumn{1}{c}{16.6} & \multicolumn{1}{c}{14.3} & \textbf{12.5} & \multicolumn{1}{c|}{5733} & \multicolumn{1}{c|}{5289} & \textbf{4826} & \multicolumn{1}{c|}{2.24} & \multicolumn{1}{c|}{1.28} & \textbf{0.35} \\ 
& \textbf{Average}        & \multicolumn{1}{c}{\textbf{\makecell[c]{26.5 \\ \textbf{(+49.7\%)}}}} & \multicolumn{1}{c}{\textbf{\makecell[c]{24.1 \\ \textbf{(+36.2\%)}}}} & \textbf{17.7} & \multicolumn{1}{c|}{\textbf{\makecell[c]{6779 \\ \textbf{(+53.0\%)}}}} & \multicolumn{1}{c|}{\textbf{\makecell[c]{6318 \\ \textbf{(+42.6\%)}}}} & \textbf{4431} & \multicolumn{1}{c|}{\textbf{\makecell[c]{3.56 \\ \textbf{(5.4$\times$)}}}} & \multicolumn{1}{c|}{\textbf{\makecell[c]{2.06 \\ \textbf{(3.12$\times$)}}}} & \textbf{0.66} \\ \hline
\multirow{6}{*}{\makecell[c]{NVIDIA \\ Jetson TX2}} & ResNet-18   & \multicolumn{1}{c}{18.9} & \multicolumn{1}{c}{14.6} & \textbf{10.8} & \multicolumn{1}{c|}{6897} & \multicolumn{1}{c|}{6438} & \textbf{5278} & \multicolumn{1}{c|}{3.14} & \multicolumn{1}{c|}{1.87} & \textbf{0.48} \\
& Inception-v4           & \multicolumn{1}{c}{16.3} & \multicolumn{1}{c}{14.9} & \textbf{12.1} & \multicolumn{1}{c|}{7018} & \multicolumn{1}{c|}{6381} & \textbf{5469} & \multicolumn{1}{c|}{2.36} & \multicolumn{1}{c|}{1.65} & \textbf{0.49} \\
& MobileNet-v2           & \multicolumn{1}{c}{21.6} & \multicolumn{1}{c}{18.4} & \textbf{14.9} & \multicolumn{1}{c|}{8248} & \multicolumn{1}{c|}{7456} & \textbf{6597} & \multicolumn{1}{c|}{3.71} & \multicolumn{1}{c|}{2.59} & \textbf{1.35} \\
& YOLOv3-Tiny            & \multicolumn{1}{c}{19.4} & \multicolumn{1}{c}{15.1} & \textbf{12.7} & \multicolumn{1}{c|}{6732} & \multicolumn{1}{c|}{6279} & \textbf{5367} & \multicolumn{1}{c|}{1.95} & \multicolumn{1}{c|}{1.46} & \textbf{0.82} \\
& RetinaNet              & \multicolumn{1}{c}{27.6} & \multicolumn{1}{c}{21.5} & \textbf{16.2} & \multicolumn{1}{c|}{9546} & \multicolumn{1}{c|}{8948} & \textbf{7294} & \multicolumn{1}{c|}{3.36} & \multicolumn{1}{c|}{2.08} & \textbf{1.23} \\
& DeepSpeech             & \multicolumn{1}{c}{12.4} & \multicolumn{1}{c}{10.7} & \textbf{8.3} & \multicolumn{1}{c|}{6754} & \multicolumn{1}{c|}{6017} & \textbf{5309} & \multicolumn{1}{c|}{2.46} & \multicolumn{1}{c|}{1.61} & \textbf{0.27} \\
& \textbf{Average}       & \multicolumn{1}{c}{\textbf{\makecell[c]{19.4 \\ \textbf{(+55.2\%)}}}} & \multicolumn{1}{c}{\textbf{\makecell[c]{15.9 \\ \textbf{(+27.2\%)}}}} & \textbf{12.5} & \multicolumn{1}{c|}{\textbf{\makecell[c]{7533 \\ \textbf{(+30.0\%)}}}} & \multicolumn{1}{c|}{\textbf{\makecell[c]{6920 \\ \textbf{(+17.6\%)}}}} & \textbf{5886} & \multicolumn{1}{c|}{\textbf{\makecell[c]{2.83 \\ \textbf{(3.68$\times$}}}} & \multicolumn{1}{c|}{\textbf{\makecell[c]{1.88 \\ \textbf{(2.44$\times$)}}}} & \textbf{0.77} \\ \hline
\end{tabular}
\end{table*}

\begin{table*}
\setlength{\abovecaptionskip}{0pt}
\setlength{\belowcaptionskip}{0pt}
    \caption{Evaluation of scalability on ImageNet-2012 dataset.}
    \label{scale_Image}
    \centering
    \begin{tabular}{c|c|ccc|ccc|ccc} \hline
\multirow{2}{*}{Edge Device} & \multirow{2}{*}{Model} & \multicolumn{3}{c|}{End-to-end latency (ms)} & \multicolumn{3}{c|}{Energy consumption (mJ)} & \multicolumn{3}{c}{Accuracy loss (\%)}  \\  
            &           & \multicolumn{1}{c}{AppealNet} & \multicolumn{1}{c}{DRLDO} & \multicolumn{1}{c|}{\textbf{DVFO}} & \multicolumn{1}{c}{AppealNet} & \multicolumn{1}{c}{DRLDO} & \multicolumn{1}{c|}{\textbf{DVFO}} & \multicolumn{1}{c}{AppealNet} & \multicolumn{1}{c}{DRLDO} & \multicolumn{1}{c}{\textbf{DVFO}} \\ \cline{1-11}
\multirow{6}{*}{\makecell[c]{NVIDIA \\ Jetson Nano}} & ResNet-18  & \multicolumn{1}{c}{28.5} & \multicolumn{1}{c}{26.7} & \textbf{16.9} & \multicolumn{1}{c|}{6534} & \multicolumn{1}{c|}{6047} & \textbf{5296} & \multicolumn{1}{c|}{3.68} & \multicolumn{1}{c|}{2.25} & \textbf{0.62} \\
& Inception-v4           & \multicolumn{1}{c}{32.6} & \multicolumn{1}{c}{30.8} & \textbf{19.8} & \multicolumn{1}{c|}{7397} & \multicolumn{1}{c|}{6854} & \textbf{5716} & \multicolumn{1}{c|}{3.06} & \multicolumn{1}{c|}{1.75} & \textbf{0.79} \\
& MobileNet-v2           & \multicolumn{1}{c}{41.6} & \multicolumn{1}{c}{36.5} & \textbf{22.1} & \multicolumn{1}{c|}{7528} & \multicolumn{1}{c|}{7112} & \textbf{6048} & \multicolumn{1}{c|}{2.13} & \multicolumn{1}{c|}{1.23} & \textbf{1.04} \\
& YOLOv3-Tiny            & \multicolumn{1}{c}{35.9} & \multicolumn{1}{c}{31.7} & \textbf{23.4} & \multicolumn{1}{c|}{7495} & \multicolumn{1}{c|}{7026} & \textbf{6208} & \multicolumn{1}{c|}{4.61} & \multicolumn{1}{c|}{3.15} & \textbf{1.14} \\
& RetinaNet              & \multicolumn{1}{c}{43.8} & \multicolumn{1}{c}{37.6} & \textbf{28.4} & \multicolumn{1}{c|}{9026} & \multicolumn{1}{c|}{7982} & \textbf{6754} & \multicolumn{1}{c|}{3.86} & \multicolumn{1}{c|}{2.74} & \textbf{1.28} \\
& DeepSpeech             & \multicolumn{1}{c}{20.7} & \multicolumn{1}{c}{17.3} & \textbf{13.8} & \multicolumn{1}{c|}{6284} & \multicolumn{1}{c|}{5749} & \textbf{5217} & \multicolumn{1}{c|}{2.46} & \multicolumn{1}{c|}{1.79} & \textbf{0.43} \\
& \textbf{Average}       & \multicolumn{1}{c}{\textbf{\makecell[c]{33.9 \\ \textbf{(+63.8\%)}}}} & \multicolumn{1}{c}{\textbf{\makecell[c]{30.1 \\ \textbf{(+45.4\%)}}}} & \textbf{20.7} & \multicolumn{1}{c|}{\textbf{\makecell[c]{7377 \\ \textbf{(+25.6\%)}}}} & \multicolumn{1}{c|}{\textbf{\makecell[c]{6795 \\ \textbf{(+15.7\%)}}}} & \textbf{5873} & \multicolumn{1}{c|}{\textbf{\makecell[c]{3.30 \\ \textbf{(3.75$\times$)}}}} & \multicolumn{1}{c|}{\textbf{\makecell[c]{1.82 \\ \textbf{(2.07$\times$)}}}} & \textbf{0.88} \\ \hline
\multirow{6}{*}{\makecell[c]{NVIDIA \\ Jetson TX2}} & ResNet-18 & \multicolumn{1}{c}{21.6} & \multicolumn{1}{c}{17.3} & \textbf{13.4} & \multicolumn{1}{c|}{7598} & \multicolumn{1}{c|}{7164} & \textbf{6025} & \multicolumn{1}{c|}{2.95} & \multicolumn{1}{c|}{1.37} & \textbf{0.74} \\
& Inception-v4           & \multicolumn{1}{c}{23.9} & \multicolumn{1}{c}{20.8} & \textbf{14.7} & \multicolumn{1}{c|}{7456} & \multicolumn{1}{c|}{6741} & \textbf{5876} & \multicolumn{1}{c|}{4.05} & \multicolumn{1}{c|}{2.82} & \textbf{1.24} \\
& MobileNet-v2           & \multicolumn{1}{c}{24.8} & \multicolumn{1}{c}{22.7} & \textbf{16.5} & \multicolumn{1}{c|}{8946} & \multicolumn{1}{c|}{8294} & \textbf{6732} & \multicolumn{1}{c|}{4.51} & \multicolumn{1}{c|}{3.28} & \textbf{1.42} \\
& YOLOv3-Tiny            & \multicolumn{1}{c}{23.8} & \multicolumn{1}{c}{20.1} & \textbf{14.5} & \multicolumn{1}{c|}{7546} & \multicolumn{1}{c|}{6853} & \textbf{5794} & \multicolumn{1}{c|}{2.74} & \multicolumn{1}{c|}{1.44} & \textbf{0.82} \\
& RetinaNet              & \multicolumn{1}{c}{30.5} & \multicolumn{1}{c}{24.2} & \textbf{20.8} & \multicolumn{1}{c|}{11542} & \multicolumn{1}{c|}{9528} & \textbf{8013} & \multicolumn{1}{c|}{4.27} & \multicolumn{1}{c|}{2.47} & \textbf{1.31} \\
& DeepSpeech             & \multicolumn{1}{c}{14.2} & \multicolumn{1}{c}{12.2} & \textbf{9.7} & \multicolumn{1}{c|}{7865} & \multicolumn{1}{c|}{7241} & \textbf{5946} & \multicolumn{1}{c|}{2.69} & \multicolumn{1}{c|}{1.53} & \textbf{0.34} \\
& \textbf{Average}       & \multicolumn{1}{c}{{\textbf{\makecell[c]{23.1 \\ \textbf{(+55.0\%)}}}}} & \multicolumn{1}{c}{\textbf{\makecell[c]{19.5 \\ \textbf{(+30.9\%)}}}} & \textbf{14.9} & \multicolumn{1}{c|}{\textbf{\makecell[c]{8492 \\ \textbf{(+32.7\%)}}}} & \multicolumn{1}{c|}{\textbf{\makecell[c]{7637 \\ \textbf{(+19.4\%)}}}} & \textbf{6398} & \multicolumn{1}{c|}{\textbf{\makecell[c]{3.54 \\ \textbf{(3.61$\times$)}}}} & \multicolumn{1}{c|}{\textbf{\makecell[c]{2.15 \\ \textbf{(2.19$\times$)}}}} & \textbf{0.98} \\ \hline
\end{tabular}
\end{table*}

\section{Related work}\label{Related}
\subsection{Learning-based DVFS}
Prior work~\cite{chen2022quality,panda2022energy,zhang2017energy,ul2015hybrid,liu2021cartad,zhang2018double,huang2019autonomous,yeganeh2020ring} has proposed a series of deep reinforcement learning-based DVFS techniques to reduce energy consumption. 
For instance, DRL quality optimizer~\cite{chen2022quality} combines deep reinforcement learning-based DVFS technology with LSTM-based selectors to reduce end-to-end latency and improve quality of service (QoS).
QL-HDS~\cite{zhang2017energy} combines Q-learning with stacked auto-encoder, and proposes a hybrid DVFS energy-saving scheduling scheme based on Q-learning.
DQL-EES~\cite{zhang2018double} and Double-Q governor~\cite{huang2019autonomous} leverage double-Q learning-based DVFS technology that dynamically scale computing frequency to achieve efficient energy-saving.
Hybrid DVFS~\cite{ul2015hybrid} considers heterogeneous workloads, dynamic relaxation and power constraints, which utilizes reinforcement learning-based hybrid DVFS technology to achieve energy-saving.
CARTAD~\cite{liu2021cartad} leverages reinforcement learning-based task scheduling and DVFS to jointly optimize end-to-end latency and temperature on multi-core CPUs systems.
Ring-DVFS~\cite{yeganeh2020ring} proposes an enhanced reinforcement learning-based DVFS technique to reduce power consumption on multi-core CPU systems.

Most related to our work is DRLDO~\cite{panda2022energy}, a data offloading scheme that combines DRL and DVFS to reduce the energy consumption of IoT devices.
However, the above-mentioned DRL-based DVFS approaches including~\cite{panda2022energy} only optimize the CPU frequency of edge devices, ignoring the impact of GPU and memory frequencies on energy consumption.
Moreover, DRLDO~\cite{panda2022energy} offloads uncompressed raw data to cloud servers, causing system instability and bandwidth bottlenecks.
In this work, we introduce DVFS into edge-cloud collaborative architecture, i.e., using DVFS for DNN feature maps offloading to further improve energy consumption performance on edge devices.
In addition, we utilize the attention mechanism to efficiently compress the original DNN feature maps, reducing the transmission delay of compressed DNN feature maps to be offloaded while maintaining accuracy.

\subsection{Edge-cloud collaborative DNN model inference}
Since edge devices are usually resource-constrained, it is necessary to utilize cloud servers with abundant computing resources for edge-cloud collaborative inference to reduce end-to-end latency. 
Existing studies~\cite{laskaridis2020spinn,hu2020starfish} have been revealed that transmission of DNN feature maps is a major network bottleneck for offloading. 
Therefore, prior work~\cite{yao2020deep,hu2020starfish,huang2022real,laskaridis2020spinn} has proposed various collaborative inference methods that combine a series of compression techniques to reduce the transmission of DNN feature maps.
For instance, DeepCOD~\cite{yao2020deep} and Starfish~\cite{hu2020starfish} designs efficient encoders and decoders based on compressed sensing theory and application-specific codecs, respectively, and then offloads the compressed data from the local to the edge server, thereby effectively reducing end-to-end delays.
AgileNN~\cite{huang2022real} uses attention mechanism to identify the importance of DNN feature maps, and it reduces the end-to-end latency by offloading a large number of less-important compressed features to remote.
SPINN~\cite{laskaridis2020spinn} achieves the progressive inference for edge-cloud collaboration by placing multiple early-exit points in the neural network, which not only considers the resource-constrained local devices, but also takes into account the instability and communication costs of the cloud.

In addition, \cite{yang2021towards} proposes a pipelined scheme for collaborative inference on a heterogeneous IoT edge cluster to reduce redundant calculation and communication overhead in order to maximize the throughput.
DCCI~\cite{hu2023content} and AppealNet~\cite{li2021appealnet} perform binary offloading (local inference only or full offloading to the cloud servers) on the input data based on the hard-case discriminator.
ELF~\cite{zhang2021elf} splits a single video frame and offloads the segmented local video frames to multiple edge servers, which accelerates parallel inference for high-resolution vision models.
CNNPC~\cite{yang2022cnnpc} jointly optimizes model partitioning and compression, which significantly speeds up collaborative inference with the end-edge-cloud computing paradigm.
However, these approaches do not incorporate optimization of hardware frequency for better energy saving. 
Our approach combines the advantages of DVFS and feature maps offloading.
 
\subsection{On-device DNN model inference}
The MLPerf Mobile Inference Benchmark~\cite{janapa2022mlperf} reveals impressive progress in on-device inference~\cite{wang2021asymo}, benefiting from the synergistic impact of increasing performance at the edge, highly flexible lightweight models, and efficient deep learning frameworks. 
We highlight that previous work~\cite{guo2021mistify,bai2022automated,pansare2022learning,wang2022towards,zhu2023enode,ling2022blastnet,zhang2023pos,wang2021asymo} can also achieve effective energy saving and low end-to-end latency only through efficient on-device inference, except for edge-cloud collaborative inference.
Mistify~\cite{guo2021mistify} and NeuralUCB~\cite{bai2022automated} studied the automated customization for on-device DL inference, which reduce DNN manual porting time and improve quality of experience (QoE) by automatically porting cloud-based compressed models to edge devices and online learning algorithms, respectively.
MEmCom~\cite{pansare2022learning} significantly improves the on-device inference performance of recommendation models by using a model compression technique based on multi-embedding compression.
DeiT-Tiny~\cite{wang2022towards} is the first empirical study on efficient on-device inference for visual transformer, reducing the end-to-end latency by removing redundant attention heads and forward neural network layers.
eNODE~\cite{zhu2023enode} achieves efficient on-device inference for neural differential equations (NODEs) by architecture-algorithm co-design.
BlastNet~\cite{ling2022blastnet} leverages dual-block based fine-grained dynamic scheduling to enable on-device real-time multi-model inference across CPU-GPU.
Similarly, POS~\cite{zhang2023pos} leverages operator granularity-oriented computational graph optimization with reinforcement learning to accelerate multi-model real-time on-device inference.
AsyMo~\cite{wang2021asymo} leverages model partitioning based on cost-model and asymmetric task scheduling for mobile CPUs to enable energy-efficient on-device inference.

As privacy security for on-device inference becomes increasingly challenging, ShadowNet~\cite{sun2022shadownet} leverages a trusted execution environment (TEE) to preserve model privacy while ensuring efficient inference.
Note that on-device inference is orthogonal to our work, which can further reduce end-to-end latency and energy consumption.

\section{Conclusion}\label{conclusion}
In this work, we propose DVFO, an DVFS enabled learning-based energy-efficient collaborative inference framework that co-optimizes the CPU, GPU, and memory frequencies of edge devices, as well as the proportion of feature maps to be offloaded to cloud servers. 
We apply concurrent control mechanism named \emph{thinking-while-moving} in learning-based optimization, and design an importance-based feature maps offloading scheme by leveraging a \emph{spatial-channel attention} mechanism, to accelerate convergence and alleviate edge-cloud network bottlenecks, respectively.
Extensive evaluations on widely-deployed DNN models with three domain-specific and three heterogeneous edge devices show that DVFO significantly outperforms existing offloading schemes in terms of energy consumption and end-to-end latency, while maintaining the average accuracy loss within 1\%.

{\small
\bibliographystyle{IEEEtran}
\bibliography{ref}
}
\vspace{-15mm}
\begin{IEEEbiography}[{\includegraphics[width=1in,height=1.25in,clip,keepaspectratio]{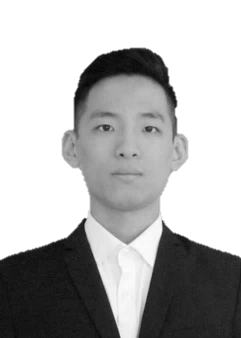}}]
{Ziyang Zhang}
received the MS degrees in the School of Electronic Information and Optical Engineering, Nankai University, Tianjin, China, in 2020.
He is currently working toward the PhD degree in the School of Computer Science and Technology, Harbin Institute of Technology (HIT), Harbin, China.
His research interests include edge computing, machine learning system, and deep learning.
\end{IEEEbiography}
\vspace{-15mm}
\begin{IEEEbiography}[{\includegraphics[width=1in,height=1.25in,clip,keepaspectratio]{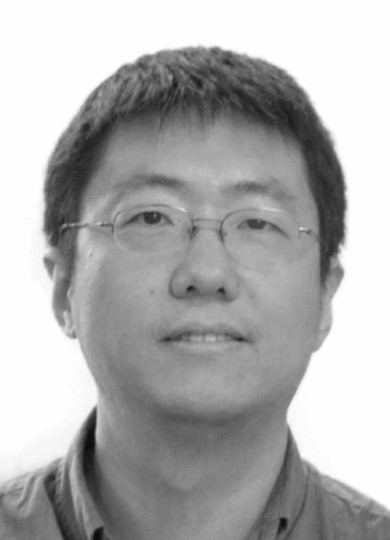}}]
{Yang Zhao}
received the BS degree (2003) in electrical engineering from Shandong University, the MS degree (2006) in electrical engineering from the Beijing University of Aeronautics and Astronautics, and the PhD degree (2012) in electrical and computer engineering from the University of Utah. 
He was a lead research engineer at GE Global Research between 2013 and 2021. Since 2021, he has been at Harbin Institute of Technology, Shenzhen, where he is a research professor in the International Research Institute for Artificial Intelligence. 
His research interests include wireless sensing, edge computing and cyber physical systems. 
He is a senior member of the IEEE. 
\end{IEEEbiography}
\vspace{-15mm}
\begin{IEEEbiography}[{\includegraphics[width=1in,height=1.25in,clip,keepaspectratio]{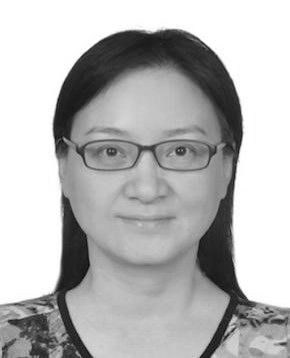}}]
{Huan Li}
obtained her PhD degree in Computer Science from the University of Massachusetts at Amherst, USA in 2006. 
Her current research interests include AIoT, Edge intelligence, distributed real-time systems, and data science. 
She has served as program committee member for numerous international conferences including IEEE RTAS, ICDCS, RTCSA, etc. 
She is now a senior member of IEEE. 
\end{IEEEbiography}
\vspace{-15mm}
\begin{IEEEbiography}[{\includegraphics[width=1in,height=1.25in,clip,keepaspectratio]{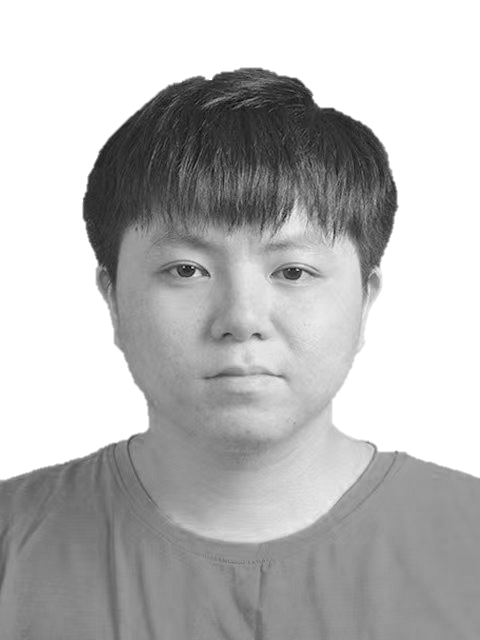}}]
{Changyao Lin}
received the BS and MS degrees in the School of Computer Science and Technology, Harbin Institute of Technology (HIT), Harbin, China, in 2020 and 2022, respectively. 
He is currently working toward the PhD degree at HIT. 
His research interests include edge computing, distributed system, and deep learning.
\end{IEEEbiography} 
\vspace{-15mm}
\begin{IEEEbiography}[{\includegraphics[width=1in,height=1.25in,clip,keepaspectratio]{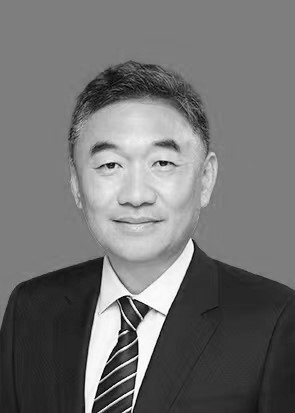}}]
{Jie Liu}
is a Chair Professor at Harbin Institute of Technology Shenzhen (HIT Shenzhen), China and the Dean of its AI Research Institute. 
Before joining HIT, he spent 18 years at Xerox PARC and Microsoft. 
He was a Principal Research Manager at Microsoft Research, Redmond and a partner of the company. 
His research interests are Cyber-Physical Systems, AI for IoT, and energy-efficient computing. 
He received IEEE TCCPS Distinguished Leadership Award and 6 Best Paper Awards from top conferences. 
He is an IEEE Fellow and an ACM Distinguished Scientist, and founding Chair of ACM SIGBED China.
\end{IEEEbiography}

\end{document}